\documentclass[lettersize,journal]{IEEEtran}

\usepackage{amsmath,amsfonts}
\usepackage{algorithmic}
\usepackage{algorithm}
\usepackage{array}
\usepackage{textcomp}
\usepackage{stfloats}
\usepackage{url}
\usepackage{verbatim}
\usepackage{graphicx}
\usepackage{diagbox} 
\usepackage{cite}
\usepackage{capt-of}
\usepackage{cuted}
\usepackage{pifont} 
\usepackage[colorlinks=true, linkcolor=blue, citecolor=blue, urlcolor=blue]{hyperref}
\newcommand{\cmark}{\ding{51}} 
\newcommand{\xmark}{\ding{55}} 

\usepackage{subcaption}

\begin{document}

\title{Automated Modeling Method for Pathloss Model Discovery}
\author{Ahmad Anaqreh, Shih-Kai Chou, Bla\v{z} Bertalani\v{c}, Mihael Mohor\v{c}i\v{c}, Thomas Lagkas, Carolina Fortuna

\thanks{Ahmad Anaqreh was with Jožef Stefan Institute and currently with Department of Computer Science, Faculty of Information Technology, Applied Science Private University(ASU), Amman, Jordan (email: a\_alanaqreh@asu.edu.jo)\\
Shih-Kai Chou, Bla\v{z} Bertalani\v{c}, Mihael Mohor\v{c}i\v{c}, and Carolina Fortuna are with Jožef Stefan Institute, Jamova cesta 39, 1000 Ljubljana, Slovenia  (email: \{shih-kai.chou; blaz.bertalanic; miha.mohorcic; carolina.fortuna\}@ijs.si)\\
Thomas Lagkas is with Department of Informatics, Democritus University of Thrace, Greece  (email: tlagkas@cs.duth.gr)

}}

\markboth{}
{Shell \MakeLowercase{\textit{et al.}}: A Sample Article Using IEEEtran.cls for IEEE Journals} 


\maketitle
\begin{abstract}
Modeling propagation is the cornerstone for designing and optimizing next-generation wireless systems, with a particular emphasis on 5G and beyond technologies, such as reconfigurable intelligent surfaces, terahertz (THz) communications, and beam steering in massive multi-input multi-output (mMIMO) systems. Traditional modeling methods have long relied on statistic-based techniques to characterize propagation behavior across different environments. With the expansion of wireless communication systems, there is a growing demand for methods that guarantee the accuracy and explicit functional interpretability of modeling. Artificial intelligence (AI)-based techniques, in particular, are increasingly being adopted to overcome this challenge. Inspired by recent advancements in AI, this paper proposes a novel approach that accelerates the discovery of pathloss (PL) models while maintaining interpretability, referring to explicit functional interpretability in the remainder of the paper unless otherwise stated. The proposed method automates the formulation, evaluation, and refinement of the model, facilitating the discovery of the model. We examine two techniques: one based on Deep Symbolic Regression (DSR), offering interpretability, and the second based on Kolmogorov-Arnold Networks (KANs), providing two levels of interpretability. Both approaches are evaluated on two synthetic and two real-world datasets. Our results show that KANs achieve the coefficient of determination value $R^2$ close to 1 with minimal prediction error, while DSR generates compact models with moderate accuracy. Moreover, on the selected examples, we demonstrate that automated methods outperform traditional methods, achieving up to 75\% reduction in prediction errors, offering accurate and functionally interpretable and analyzable solutions with potential to increase the efficiency of discovering next-generation pathloss models.
\end{abstract}


\section{Introduction}

\textit{Emerging AI for Scientific Discovery:} Discovering accurate and interpretable scientific formulas that align with established theories is a fundamental goal in science. Traditionally, scientists formulate a hypothesis, derive formulas manually to test it, relying exclusively on theoretical knowledge or statistical characteristics, and subsequently verify its accuracy through simulations and experiments. However, this manual approach, which excludes experimental data during the discovery phase, can be time-consuming. Automated methods have emerged to overcome this limitation by incorporating data directly into the discovery process in several disciplines ~\cite{Jumper,Schmidt}. Thus, accurate and interpretable formulas are generated, significantly improving both the speed and quality of the modeling process compared to traditional methods. For instance, recent work in high-energy physics \cite{nour2024} has demonstrated that symbolic regression (SR) can derive models from experimentally measured data that closely resemble the Tsallis distribution, a well-known statistical distribution. Another application lies in improving localization in various domains such as wireless communications, robotics \cite{wang2025}, and the Internet of Things~\cite{shahbazian}.

\begin{figure*}[h!]
    \centering
    \includegraphics[width=1\textwidth]{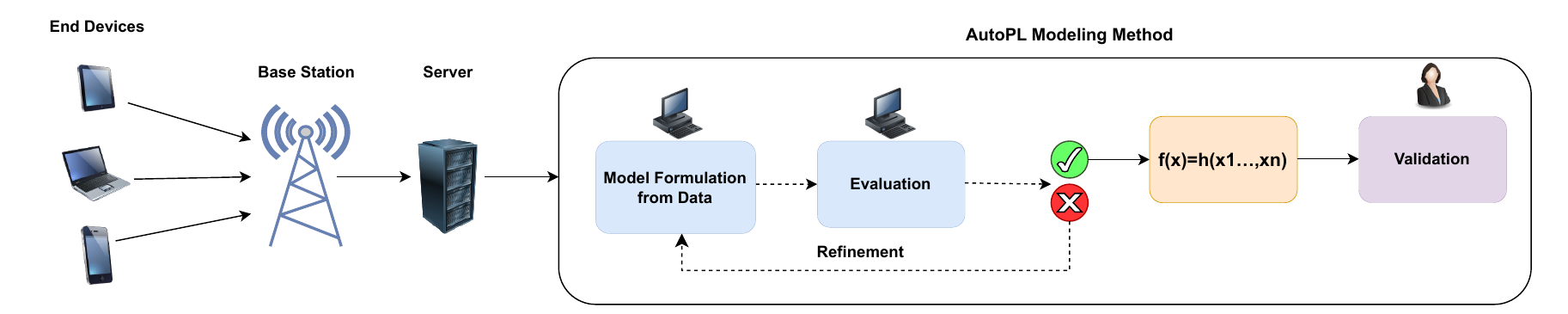} 
    \caption{The flow diagram of AutoPL modeling method. }
    \label{fig:modeling_workflow}
\end{figure*}

Unlike AI scientists \cite{lu2024ai}, who aim to even replace humans in the scientific discovery process, automated methods aim to support the human user to be able to discover interpretable relationships directly from data, without relying on predefined models, by seeking functional representations that capture the underlying patterns \cite{makke2024interpretable}. By leveraging tools such as evolutionary strategies~\cite{koza1994} and neural network architectures~\cite{Petersen}, these methods can efficiently model complex relationships. 

\textit{AI for PL Modeling:} Understanding specific electromagnetic propagation models is paramount for wireless networks that have been engineered over time by humans and enable our everyday communication needs. Our ability to build and understand these networks hinges on understanding and modeling how signals are attenuated when propagating through space in realistic environments \cite{Phillips}. Pathloss (PL) models describe the attenuation behavior of signal propagation in various wireless communication systems, incorporating essential factors such as frequency, distance, and the presence of various obstacles along the path \cite{Sun}. Over the years, a variety of methods have been proposed for modeling PL, including empirical, deterministic, and machine learning-based approaches, each with its strengths and limitations \cite {Phillips,Ethier}. The traditional empirical models, such as the widely used Alpha-Beta-Gamma (ABG) and Close-In (CI) models, rely on rigid, predefined forms. This structural inflexibility limits their ability to capture the complex environmental effects, including the dual-slope attenuation characteristic of urban streets~\cite{slope_ref} or the molecular absorption peaks inherent to Terahertz (THz) communications~\cite{molecular_ref}.

\textit{Need for Functional Interpretability:} In recent years, such models have been modeled with deep neural networks (DNNs) \cite{Zeng}, in some cases showing better results than some existing analytical models and classical machine learning-based models \cite{Ethier}. However, models based on DNNs are inherently difficult to interpret and explain \cite{dwivedi2023explainable}, which hinders the ability to teach humans the principles and intuitions behind models. Unlike the traditional, physics-based methods, DNNs do not provide an explicit mathematical expression to describe the relationship between parameters (e.g., frequency, distance) and the signal attenuation. This lack of interpretability, referring to explicit functional interpretability in the remainder of the paper unless otherwise stated, makes it difficult for future engineers or researchers to validate the model against physical laws or derive generalizable insights for network planning, essentially trading physical understanding for numerical accuracy.



To bridge the aforementioned gap, we propose AutoPL, a novel approach that accelerates the discovery of PL models. Unlike traditional methods where experts manually formulate and refine models based on established theories, AutoPL automates the symbolic discovery process to retrieve interpretable mathematical expressions directly from data (see Figure~\ref{fig:modeling_workflow}). Motivated by the observation that PL models often involve logarithmic functions~\cite{Phillips}, which can present challenges for certain discovery techniques~\cite{makke2024interpretable}, we investigate this data-driven alternative to ensure \textit{explicit functional} interpretability. In this context, interpretability refers to the availability of explicit mathematical expressions and transparent parameter dependencies, rather than a direct interpretation of underlying electromagnetic propagation mechanisms. The contributions of this paper are outlined as follows.


\begin{enumerate}
    \item We propose a new PL model discovery approach (AutoPL) that automates the model formulation, evaluation, and refinement, facilitating model discovery. In particular, we consider Deep Symbolic Regression (DSR), offering explicit functional interpretability, and  Kolmogorov-Arnold Networks (KANs), providing two flavors of explicit functional interpretability, as candidate automation techniques.
    \item Our results show the feasibility of the proposed automation method in two synthetic datasets and two real-world datasets. More specifically, we show that KANs achieve \( R^2 \) values close to 1 with minimal prediction error, while DSR generates compact models with moderate accuracy. 
    \item On the selected examples, we demonstrate that automated methods outperform traditional methods, achieving up to 75\% reduction in prediction errors, offering accurate and explainable solutions with the potential to increase the efficiency of discovering PL models for next-generation wireless communication systems.
    \item We provide the automatic KAN-based model discovery workflow for all scenarios as open source\footnote{
    https://github.com/sensorlab/AutoPL}. 
\end{enumerate}

The paper is organized as follows. Section \ref{sec:relatedwork} provides a review of the relevant literature, followed by a description of the problem in Section \ref{sec:probstat}. Section \ref{sec:met} details the methods employed to address the problem, while Section \ref{sec:methodology} details the training data and training methodologies. Section \ref{sec:results} presents the results. Finally, Section \ref{sec:con} concludes the paper and discusses potential future research directions.

\section{Related work}
\label{sec:relatedwork}
In this section, we first summarize related work on automated scientific discovery, followed by related work on AI for PL modeling.

\subsection{Automating Scientific Discovery}
Perhaps the most recent trend in automating scientific discovery is to develop AI systems that can automate every step of the process, including idea generation, full experiment iteration, and paper writing \cite{lu2024ai}. Slightly more developed, but still in the very early stages of research, various approaches to automating hypothesis generation and verifications have been proposed \cite{cory2024evolving}.

SR is one of the earliest and most widely used modeling techniques. A breakthrough came with the introduction of genetic programming (GP), which refines mathematical expressions using evolutionary operations such as mutation, crossover, and selection\cite{koza1994}. GP has been extensively used in SR and remains a dominant approach due to its ability to discover compact expressions \cite{Fortin}. However, GP has significant computational challenges, particularly when searching through large, high-dimensional function spaces \cite{Kronberger}. As the complexity of mathematical expressions increases, the search space expands exponentially, resulting in challenges such as local optima, slow convergence, and insufficient global search guidance. These difficulties have driven the development of hybrid approaches that integrate deep learning (DL) and reinforcement learning (RL) with SR, with the aim of improving efficiency and exploration.

Mundhenk et al.\cite{Mundhenk} introduced a neural-guided genetic programming (NGGP) approach, where a recurrent neural network generates initial populations for GP. Unlike standard GP, NGGP uses a trained neural network to initialize the GP search with higher-quality candidate expressions, leading to faster convergence. A closely related approach, Genetic Expert-Guided Learning (GEGL) \cite{Ahn}, combines priority queues and GP to select high-quality expressions for RL. However, GEGL uses only one evolutionary step per training iteration, which limits the depth of genetic refinement.

Other hybrid models integrate RL to adjust mutation probabilities, guide selection criteria, or dynamically optimize hyperparameters \cite{Igel,Topchy}. These approaches introduce mechanisms to bias evolutionary search toward higher-reward expressions, helping GP escape local optima while maintaining structural consistency in generated formulas. Another class of hybrid approaches interchanges populations between GP and neural models. Pourchot et al. \cite{Pourchot} and Khadka et al. \cite{Khadka} applied population interchange in deep reinforcement learning tasks such as robotic control. These approaches demonstrate that GP-generated expressions can be used to refine neural network policies, while neural models can improve the diversity of GP populations.

Recent advances in DL have introduced alternative methods for SR. Kusner et al. \cite{Kusner} proposed GrammarVAE, which uses a variational autoencoder to model symbolic expressions as latent-space representations while enforcing syntactic constraints. However, the method struggles with recovering exact expressions and often generates invalid mathematical structures. The approach proposed in \cite{Sahoo}  incorporates SR into a neural network architecture, where activation functions represent symbolic operators. This approach requires strong assumptions about the functional form of the target equation, which limit its flexibility. A major advancement came with AI Feynman \cite{Udrescu}, which combines neural networks with symbolic rule discovery to simplify expressions before applying SR. AI Feynman successfully identifies symmetries and separable functions in data, allowing it to break down complex problems into smaller, easier-to-learn subproblems. However, it does not perform a direct symbolic search, making it unsuitable for some real-world applications.

RL-based approaches have shown notable improvements, particularly in terms of accuracy and convergence speed. One of the key challenges in RL is the efficient sampling of promising expressions. Importance sampling, first introduced by Glynn \& Iglehart \cite{Glynn}, has been adapted in RL-based approaches to prioritize high-reward expressions while reducing computational cost. Another powerful optimization technique is the cross-entropy method (CEM) \cite{DeBoer}, which has been widely used in RL for sampling high-quality solutions from a probabilistic distribution. These approaches have helped address the exploration-exploitation trade-off, ensuring meaningful expressions without excessive random exploration.

KANs \cite{Liukan} represent a significant advancement and shift in neural network design by enhancing the ability to model complex data relationships while maintaining high accuracy and explicit functional interpretability. The KANs architecture plays a crucial role in achieving these benefits across various domains. Xu et al. \cite{Xu} applied KANs to time series forecasting, introducing T-KAN and MT-KAN, which demonstrated improved predictive performance by effectively capturing temporal dependencies. De Carlo et al\cite{DeCarlo} developed the Graph Kolmogorov-Arnold Network (GKAN), integrating spline-based activation functions into graph neural networks, leading to enhanced accuracy and interpretability in tasks like node classification and link prediction. Bozorgasl \& Chen \cite{Bozorgasl} proposed Wav-KAN, incorporating wavelet functions into KANs to efficiently capture both high and low frequency components, improving accuracy and robustness, particularly for multiresolution analysis. By design, KANs enable more interpretability compared to other DNNs as they learn the composition of splines that approximate the data. By looking at the learned graph and the splines at each node, one can interpret what has been learned from the data and why. More recently, they have been extended with symbolic functionality where an actual expression can be extracted from the graph by approximating the learned splines with known functions \cite{liu2024kan}. 


\subsection{AI for PL}

To date, our preliminary PL approximation study \cite{anaqreh} published in the same month with \cite{de2024analytical} are the first work to consider interpretable AI modeling of wireless networks. While \cite{anaqreh} focused on KANs and DSR with the ABG and CI models, \cite{de2024analytical} focused on symbolic regression through an evolutionary algorithm and ITU-R P.1546-5. As this paper was available as pre-print and underwent revision, KAN-based modeling has also been considered for radio map prediction \cite{11143592}. Several works on explainable AI for various aspects of 6G have been proposed, as surveyed in \cite{brik2024explainable}. In some cases, the developed models also claim interpretability, or "local" interpretability, however, depending on the work, the term refers either to a visual illustration of how much importance a model assigns to certain features or how a DL network activates, i.e., activation maps. Of course, certain classical machine learning models, also referred to as "transparent models" in \cite{brik2024explainable}, such as regression trees or linear regression, are also interpretable; however, they tend to yield less accurate models.  

With respect to AI-aided PL modeling, \cite{Zeng} proposed a meta-learning approach based on DNNs to develop a PL prediction model when only sparse measurements are available. However, its superiority compared to conventional techniques is not immediately clear. On the other hand, \cite{Ethier} showed that a fully connected network was superior to a boosted tree model and two conventional approaches in several scenarios.

\section{Problem statement}
\label{sec:probstat}

The guiding hypothesis of this work is the possibility of automating the formulation, evaluation, and refinement steps of the conventional PL model development process. By doing so, the aim is to equip scientists and engineers with more efficient tools for discovering PL models and, at the same time, maintaining human interpretability through mathematical expressions. In order to validate this hypothesis, we consider two classes of models: analytical and empirical. The two considered analytical models are $PL^{ABG}$ (Alpha-Beta-Gamma) and $PL^{CI}$ (Close-in) models~\cite{Sun2}, which are widely used in modeling PL in wireless systems. The two empirical models, herein referred to as $PL^{EI}$ and $PL^{EO}$, correspond to indoor and outdoor environments, respectively, and have been developed from measurements in ~\cite{ElChall}. As analytical baselines for corresponding PL models, we include the indoor multiwall-and-floor ($PL^{MWF}$) and outdoor free space ($PL^{FS}$) models. Based on the data generated with the analytical and baseline models, as well as the empirically measured one, we aim to automatically learn the most likely expression, i.e., formulas, that best describe the respective PL. 

Similar to established PL models such as CI and ABG, which abstract complex electromagnetic interactions into compact empirical laws, the flow proposed in Figure \ref{fig:modeling_workflow} aims to speed up the development of models that prioritize analytical usability and generalization over microscopic physical decomposition.

\subsection{ABG and CI Analytical Models}
\label{sec:abg_ci_models}
The ABG and CI models are particularly used for describing signal propagation in urban micro- and macro-cellular scenarios, incorporating frequency, distance, and shadowing factors.

The ABG model is defined by the equation:
\begin{multline}
PL^{\text{ABG}}(f, d)= 10\alpha \log_{10}\left(\frac{d}{1\,\text{m}}\right) + \beta + 10\gamma \log_{10}\left(\frac{f}{1\,\text{GHz}}\right) \\+ \chi_\sigma^{\text{ABG}},
\label{eq:abg}
\end{multline}
\noindent where \( PL^{\text{ABG}}(f, d) \) is the PL in dB, \( \alpha \) and \( \gamma \) are coefficients that represent the dependence on distance (\( d \)) and frequency (\( f \)) respectively. \( \beta \) is an optimized offset value for PL, and \( \chi_\sigma^{\text{ABG}} \)  is a random variable that represents shadow fading, which takes a random value following the Gaussian distribution with zero mean and a standard deviation of $\sigma$.

The following equation describes the CI model:
\begin{multline}
PL^{\text{CI}}(f, d) = FSPL(f, 1\,\text{m}) + 10n \log_{10}(d) + \chi_\sigma^{\text{CI}},
\label{eq:ci}
\end{multline}
\noindent where $f$ is the frequency in Hz, \( n \) is the PL exponent (PLE), \( d \) is the transmitter-receiver distance in three dimensions, \( FSPL(f, 1\,\text{m}) \) is the loss of the path in free space at 1 meter and \( \chi_\sigma^{\text{CI}} \) represents the shadow fading, described as a random variable following a Gaussian distribution with a mean of zero and a standard deviation of $\sigma$.
\begin{equation}
FSPL(f, 1\,\text{m}) = 20 \log_{10}\left(\frac{4\pi f}{c}\right), 
\end{equation}
\noindent where \( c \) is the speed of light.
\subsection{The Indoor and Outdoor Empirical Models}
\label{sec:empirical}

El Chall et al.\cite{ElChall} derived empirical PL models for LoRaWAN (Long-Range Wide Area Network) based on extensive field measurements conducted in Lebanon. Indoor experiments were carried out at Saint Joseph University, while outdoor experiments were performed in Beirut City. The collected data were analyzed using linear regression to derive PL models for both indoor and outdoor environments.

For indoor environments, they proposed the following PL model:
\begin{equation}
    PL^{EI}(d) = 10n \log_{10}(d) + PL_0 + n_w L_w + n_f^{\left( \frac{n_f + 2}{n_f + 1} - b \right)} L_f,
    \label{eq:in}
\end{equation}

\noindent where $n = 2.85$ is the PL exponent, $PL_{0} = 120.4$ is the reference PL, $b = 0.47$ is a fitting parameter, $L_{f} = 10$ and $L_{w} = 1.41$ are the attenuation factors for floors and walls, respectively.

The PL model for outdoor environments is given by:
\begin{equation}
    PL^{EO}(d) = 10n \log_{10}(d) + PL_0 + L_h \log_{10}(h_{\text{ED}}) + X_{\sigma},
    \label{eq:out}
\end{equation}

\noindent where \( L_h \) denotes the additional loss related to the antenna height \( h_{\text{ED}} \). The parameters obtained from the fitting process are: PL exponent \( n = 3.119 \), the reference PL \( PL_{0} = 140.7 \), and the antenna height loss factor \( L_h = -4.7 \), \( X_{\sigma}\) is a variable that represents shadow fading which takes a random value following the Gaussian distribution with zero mean and standard deviation of $\sigma$, where $\sigma$=9.7.

\subsection{The Baseline Multiwall-and-Floor and Free Space Models}
\label{sec:ldm} \label{sec:fsm}

To describe PL within buildings, it is essential to incorporate the additional attenuation caused by walls and floors. Therefore, the PL can be defined as:
\begin{equation}
    PL^{MWF}(d) = 10n \log_{10}\left(d\right) + PL_0 + \mathrm{WAF} + \mathrm{FAF},
    \label{eq:fw}
\end{equation}

where \( d \) is the distance between the transmitter and the receiver, \( \text{WAF} = n_w L_w \) and \( \text{FAF} = n_f L_f \) represent the wall and floor attenuation factors, which are determined by the number of walls \( n_w \) and floors \( n_f \) traversed. Based on the empirical models discussed in Section~\ref{sec:empirical}, the PL exponent is \( n = 2.85 \), with a reference PL of \( PL_0 = 120.4 \). The attenuation factors are \( L_f = 10 \) for floors and \( L_w = 1.41 \) for walls.

The Free Space model is a baseline approach used to estimate PL when the transmitter and receiver have a clear line-of-sight without any obstacles, defined as follows:
\begin{equation}
    PL^{FS}(f,d) = 20 \log_{10}(f) + 20 \log_{10}(d) + 32.44,
    \label{eq:fs}
\end{equation}
where $f$ is measured in MHz and $d$ in km.

\section{Methodologies}
\label{sec:met}
For automating the PL modeling of the models identified in Section \ref{sec:probstat}, and to further validate the working hypothesis, we identify two automated modeling methods capable of learning a symbolic representation from the data at hand: the recently introduced KANs and DSR. 

The design decision to adopt KANs and DSR as candidate techniques to enable the automation of PL learning is justified by the shortcomings of typical techniques in learning logarithmic functions \cite{makke2024interpretable}. Knowing that existing PL models often involve logarithmic functions in their expression \cite{Phillips}, KANs and DSR may have superior performance in approximating such models while being able to produce different flavors of functional interpretability. 

Overall, different methods provide varying levels of interpretability. DNNs offer \textbf{low} interpretability through feature importance or activation maps. In contrast, KANs offer a \textbf{medium} level of structural interpretability through their spline-based representations, KAN graphs and edge weights. DSR and  KANs (by using Auto-symbolic mapping) can achieve \textbf{high} functional interpretability by expressing models as explicit mathematical functions. Neither level of interpretability implies a direct mapping to individual electromagnetic effects; instead, both focus on making the learned large-scale propagation laws transparent and analyzable and eventually validated by humans as depicted in Figure \ref{fig:modeling_workflow}.

\subsection{Kolmogorov-Arnold Networks}
\label{sec:kan}
KANs have recently emerged as an alternative to multi-layer perceptron (MLP) based architectures, demonstrating strong performance on scientific tasks while maintaining interpretability. KANs have strong mathematical foundations based on the Kolmogorov-Arnold representation theorem, which states that any multivariate continuous function can be expressed as a sum of univariate functions~\cite{Liukan}. KANs and MLPs implement activation functions in fundamentally different ways: KANs have activation functions on edges, while MLPs have activation functions on nodes. This unique design allows KANs to provide a more transparent and interpretable mapping between inputs and outputs, enhancing explainability while preserving the model's ability to capture complex patterns.

The structure of a KAN can be represented as $[n_1,…,n_{L+1}]$, where $L$ denotes the total number of layers. A deeper KAN is constructed by composing $L$ layers as follows:
\begin{equation}
    Y = \text{KAN}(X) = (\Phi_L \circ \Phi_{L-1} \circ \dots \circ \Phi_1)X,
\label{eq:eq_phi_composition}
\end{equation}
Each layer of a KAN is represented by a matrix in which each entry is an activation function. If a layer contains $d_{in}$ nodes and its adjacent layer has $d_{out}$ nodes, the layer can be expressed as a $d_{in} \times d_{out}$ matrix $\Phi$ of activation functions:
\begin{equation}
\Phi = \{\phi_{q,p}\}, \; p = 1, 2, \dots, d_{in}, \; q = 1, 2, \dots, d_{out},
\end{equation}
KAN utilizes the SiLU activation function combined with B-splines to enhance expressiveness. In this setup, edges control the transformations between layers, while nodes perform simple summation operations. A B-spline of order $k$ requires $G+k$ basis functions to define the spline over a given grid. Consequently, for each input (node in a layer), evaluating a B-spline of order $k$ involves computing $G+k-1$ basis functions and performing a weighted sum with the corresponding control points:
\begin{equation}
\text{spline}(x) = \sum_{i=0}^{G+k-1} c_i B_i(x),
\end{equation}
\subsection{Deep Symbolic Regression (DSR)}
\label{sec:dsr}
At the core of DSR \cite{Petersen} are the representation of mathematical expressions as sequences, the development of an autoregressive model to generate these expressions, and the use of a risk-seeking policy gradient approach to train the model to produce more accurate and well-fitting expressions.

The sequence generator defines a parameterized distribution over mathematical expressions, \( p(\tau \mid \theta) \). The model is typically designed to ensure computational tractability of expression likelihoods for the parameters \( \theta \), allowing for backpropagation with a differentiable loss function. A common implementation is a recurrent neural network (RNN), where the likelihood of the \( i \)-th token (\( \tau_i \)) is conditionally independent of other tokens, given the initial ones (\( \tau_1, \dots, \tau_{i-1} \)). This is expressed as:
\begin{equation}
p(\tau_i \mid \tau_{j \neq i}, \theta) = p(\tau_i \mid \tau_{j < i}, \theta),
\end{equation}
The sequence generator is generally trained using RL or related techniques. From this perspective, the sequence generator works as an RL policy to be optimized. This process involves sampling a batch of \( N \) expressions \( \mathcal{T}\), evaluating each expression using a reward function \( R(\tau) \), and applying gradient descent to minimize a loss function. In this work, we employ three approaches to train the RNN.

\subsubsection{Risk-Seeking Policy Gradient (RSPG)} 
Originally proposed in \cite{Petersen}, this approach prioritizes optimizing the best-case reward rather than the average. The loss function is defined as:
\begin{equation}
\mathcal{L}(\theta) = \frac{1}{\epsilon |\mathcal{T}|} \sum_{\tau \in \mathcal{T}} \big(R(\tau) - \tilde{R}_\epsilon\big) \nabla_\theta \log p(\tau \mid \theta) \mathbf{1}_{R(\tau) > \tilde{R}_\epsilon},
\end{equation}
where \( \epsilon \) is a hyperparameter controlling the level of risk-seeking, \( \tilde{R}_\epsilon \) is the empirical \( (1 - \epsilon) \) quantile of the rewards in \( \mathcal{T} \), and \( \mathbf{1} \) is an indicator function.

\subsubsection{Vanilla Policy Gradient (VPG)} 
This method employs the REINFORCE algorithm \cite{Williams}, where training is performed over the batch \( \mathcal{T} \) with the loss function:
\begin{equation}
\mathcal{L}(\theta) = \frac{1}{|\mathcal{T}|} \sum_{\tau \in \mathcal{T}} \big(R(\tau) - b\big) \nabla_\theta \log p(\tau \mid \theta),
\end{equation}
where \( b \) is a baseline term, such as an exponentially weighted moving average (EWMA) of rewards.
\subsubsection{Priority Queue Training (PQT)} 
Proposed by Abolafia \cite{Abolafia}, this non-RL approach focuses on optimizing best-case performance. It stores samples from each batch in a maximum reward priority queue (MRPQ), and training is conducted on these stored samples using a supervised learning objective:
\begin{equation}
\mathcal{L}(\theta) = \frac{1}{k} \sum_{\tau \in \text{MRPQ}} \nabla_\theta \log p(\tau \mid \theta)    ,
\end{equation}
where \( k \) is the size of the MRPQ.

For a pre-order traversal $\tau$ and a dataset of \( (X, y) \) pairs of size \( N \), where \( X \in \mathbb{R}^n \) and \( y \in \mathbb{R} \), the normalized root-mean-square error (NRMSE) is defined as:
\begin{equation}
 \text{NRMSE}(\tau) = \frac{1}{\sigma_y} \sqrt{\frac{1}{N} \sum_{i=1}^N (y_i - f(X_i))^2},
\end{equation}
where \( f: \mathbb{R}^n \to \mathbb{R} \) is the mathematical expression instantiated from \( \tau \), and \( \sigma_y \) is the standard deviation of \( y \). Thus, the reward function is described as:
\begin{equation}
R(\tau) = \frac{1}{1 + \text{NRMSE}(\tau)},
\end{equation}
\section{Evaluation Methodologies}
\label{sec:methodology}
To ensure complete replicability of the study and clarify the conditions under which the evaluation was conducted, this section first discusses the type of data used to train the automatic discovery methods, followed by the training methodologies for each method.

\begin{figure*}[h]
    \centering
    \includegraphics[width=1\textwidth]{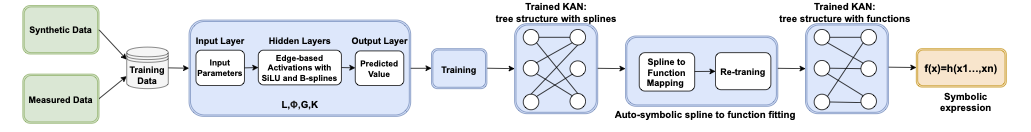} 
    \caption{The flow diagram of developing a PL model using  KANs.}
    \label{fig:kan_workflow}
\end{figure*}
\subsection{Training Data}
\subsubsection{Synthetic Data}
The dataset was generated following the approach described in \cite{anaqreh} to approximate the PL models defined by Eq.~(\ref {eq:abg}) and Eq.~(\ref{eq:ci}) in Section~\ref{sec:abg_ci_models}. The parameter ranges, summarized in Table~\ref{tab:ABG_CI_parameters}, were derived from authentic data in studies that comprehensively analyze various scenarios and frequency ranges, highlighting the applicability of $PL^{ABG}$ and $PL^{CI}$ models in various contexts \cite{Sun,Sun2,Cheng}. Specifically, for $PL^{ABG}$ the parameters $\alpha$, $\gamma$, $\beta$, $f$, $d$, and $\chi^{ABG}_\sigma$ were used as inputs, while for $PL^{CI}$ $f$, $n$, $d$, and $\chi^{CI}_\sigma$ were the inputs. The dataset consists of 1,000 instances. The frequency was converted to hertz for the $PL^{CI}$, following standard practice in the literature.

\begin{table}[h!]
\centering
\caption{The parameters and their corresponding values for $PL^{ABG}$ and $PL^{CI}$.}
\begin{tabular}{|c|c|c|}
\hline
\textbf{Parameter} & \textbf{ABG} & \textbf{CI} \\
\hline
$\boldsymbol \alpha$ & $[0.1, 2.5]$  & -        \\
\hline
$\boldsymbol \beta$   & $[-10, -1]$   & -        \\
\hline
$\boldsymbol \gamma$  & $[0, 2]$      & -        \\
\hline
$\boldsymbol f(GHz)$       & $[2, 73.5]$   & $[2, 73.5]$ \\
\hline
$\boldsymbol d(m)$       & $[1, 500]$    & $[1, 500]$  \\
\hline
$\boldsymbol \sigma$    & $[4, 12]$     & $[4, 12]$   \\
\hline
$\boldsymbol n$       & -             & $[2, 6]$    \\
\hline
\end{tabular}
\label{tab:ABG_CI_parameters}
\end{table}

\subsubsection{Empirically Measured Data}
El Chall et al. \cite{ElChall} conducted measurements in both indoor and outdoor environments. The indoor dataset includes 1,317 instances, and the outdoor dataset contains 787 instances. Input parameters for $PL^{EI}$ included $n_w$, $n_f$, $d$, and $f$ , while the $PL^{EO}$ used $h_{\mathrm{ED}}$, $d$, and $f$. The corresponding parameter ranges are presented in Table~\ref{tab:empirical_parameters}. For performance evaluation, we compared the automated methods with the models described in Section~\ref{sec:empirical} and developed by the authors in \cite{ElChall} using the real-world measurements\footnote{https://zenodo.org/records/1560654}.

\begin{table}[h!]
\centering
\caption{The parameters and their corresponding values for $PL^{EI}$ and $PL^{EO}$.}
\begin{tabular}{|c|c|c|}
\hline
\textbf{Parameter} & \textbf{EI} & \textbf{EO} \\
\hline
$\boldsymbol n_w$ & $[0, 3]$  & -        \\
\hline
$\boldsymbol n_f$   & $[1, 4]$   & -        \\
\hline
$\boldsymbol f(MHz)$       & $[868.1, 868.5]$   & $[868.1, 868.5]$ \\
\hline
$\boldsymbol d(m)$       &   $[6.47, 105.25]$  & $[27.66, 170.44]$  \\
\hline
$\boldsymbol h_{\text{ED}}(m)$       & -             & $[0.2, 3]$    \\
\hline
\end{tabular}
\label{tab:empirical_parameters}
\end{table}

\subsubsection{Analytical Baselines}
The $PL^{MWF}$ model, defined in Eq.~(\ref{eq:fw}), was used as the baseline for the indoor environment, whereas the $PL^{FS}$ model, defined in Eq.~(\ref{eq:fs}), was employed as the baseline for the outdoor scenario.

\subsection{Training of KANs}
\label{sec:kantr}
We used the original KAN implementation from the pykan library\footnote{https://github.com/KindXiaoming/pykan} and trained the KANs in a similar way as other ML models, using the 80-20 split of data as depicted in Figure \ref{fig:kan_workflow}.  During training, layers and grid sizes are configured to optimize performance. In this process, the splines and weights of the edges connecting these splines, as described in Section \ref{sec:kan}, are learned. Unlike MLPs, KANs are not very deep (i.e., usually about 3 layers with 5 being very deep) and not very wide \cite{Liukan}. Due to their relative shallowness and narrowness, once learned, they can be visualized as a tree structure with the learned activation functions overlayed on the edges and the importance of compositions represented as edge weight. We refer to this structure as tree structure with splines in the workflow illustrated in Figure \ref{fig:kan_workflow}. This alone already gives more intuition on the learned model than in the case of other neural architectures. 

As a second step, a mechanism to match the learned splines with actual known functions also exists~\cite{Liukan} in the original pykan library. The matching approach relies on a configurable library of elementary functions such as $\sin(\cdot)$, $\exp(\cdot)$, and $\log(\cdot)$. These functions are ranked based on how closely they match the spline using $R^2$ and the closest match with the simplest form is preferred. The selected function is then refined using least squares to match the spline as close as possible \cite{Liukan}.  This mechanism is implemented in the library's default $auto\_symbolic()$ functionality, which we use and is represented as the Auto-symbolic bloc in Figure \ref{fig:kan_workflow}. After the matching step, the network is retrained with the respective functions by only updating the weights of the edges resulting in the tree structure with functions from Figure~\ref{fig:kan_workflow}. This second tree structure can be re-written as a final symbolic expression. As can be seen from the figure, KANs with the auto-symbolic mapping functionality are able to provide a two-level interpretability insight: a tree with spline composition and a symbolic expression. To tune the KANs, we perform a full grid search for each model with $grids \in \{5, 8, 10, 15, 20, 30, 40, 50\}$, $steps \in \{50, 100, 200, 300\}$ and $lambda \in \{0.02, 0.002, 0.0002, 0.00002\}$. 

\begin{table}[h!]
\centering
\caption{Tuned hyperparameters for KANs.}
\begin{tabular}{|c|c|c|c|c|}
\hline
\textbf{Layers}  & \textbf{Grids} & \textbf{Steps} &\textbf{k}& $\boldsymbol{\lambda}$ \\\hline
\multicolumn{5}{|c|}{ABG model}\\
\hline
$[6,6,1]$ & 10 & 100 & 3&0.002 \\
\hline
\multicolumn{5}{|c|}{CI model}\\
\hline
$[4,4,1]$&8&300&3&0.002\\
\hline
\multicolumn{5}{|c|}{Indoor model}\\
\hline
$[4,1]$ & 5 & 100 & 3&0.0002 \\
\hline
\multicolumn{5}{|c|}{Outdoor model}\\
\hline
$[3,1]$ & 50 & 100 & 3&0.02 \\
\hline
\end{tabular}
\label{tab:KANs_hyperparams}
\end{table}

\begin{figure*}[h!]
    \centering
    \begin{subfigure}[b]{0.48\textwidth}
        \centering
        \includegraphics[width=\textwidth]{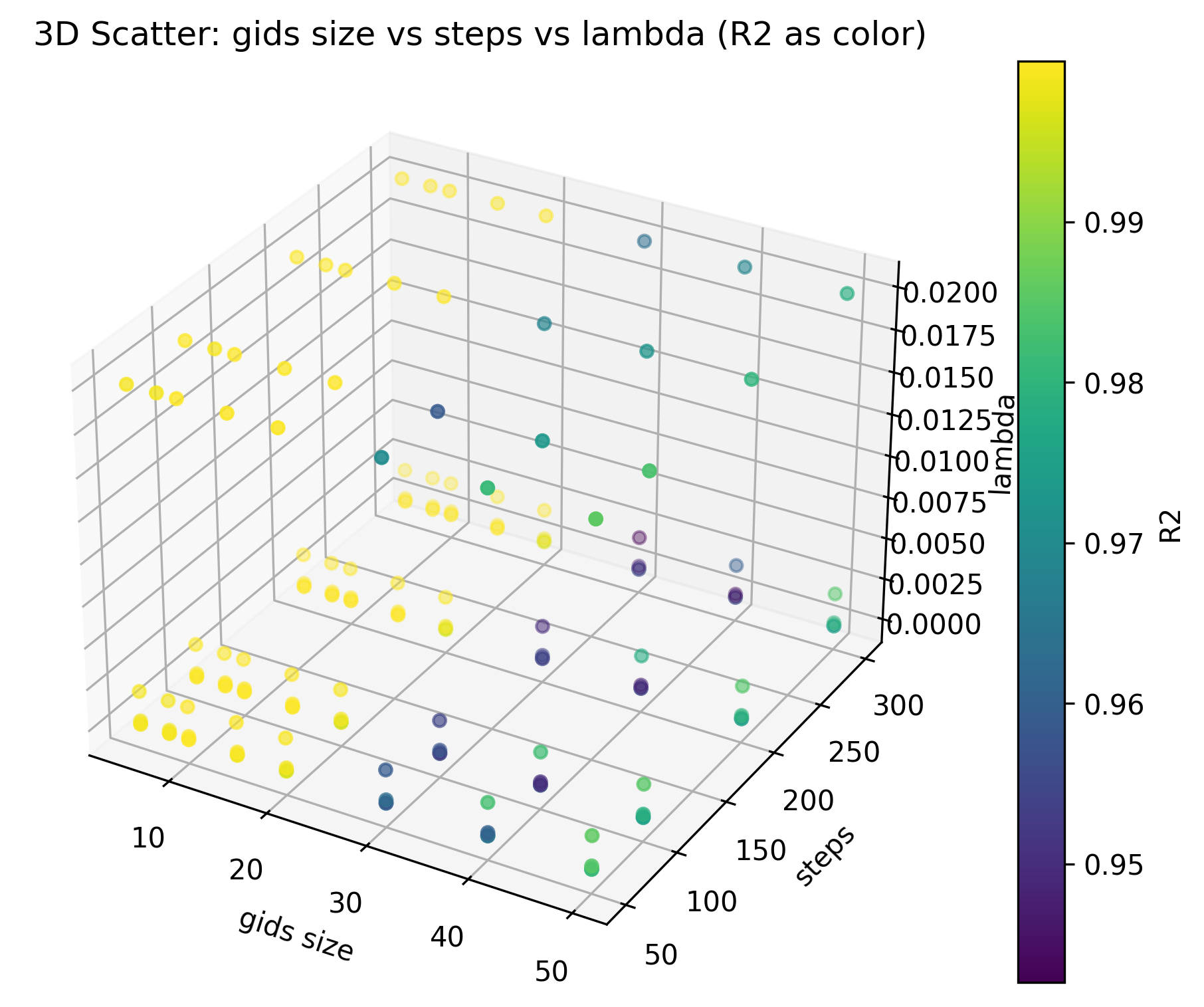}
        \caption{Parameter search for ABG.}
        \label{fig:kan_grid_r2}
    \end{subfigure}
    \hfill
    \begin{subfigure}[b]{0.48\textwidth}
        \centering
        \includegraphics[width=\textwidth]{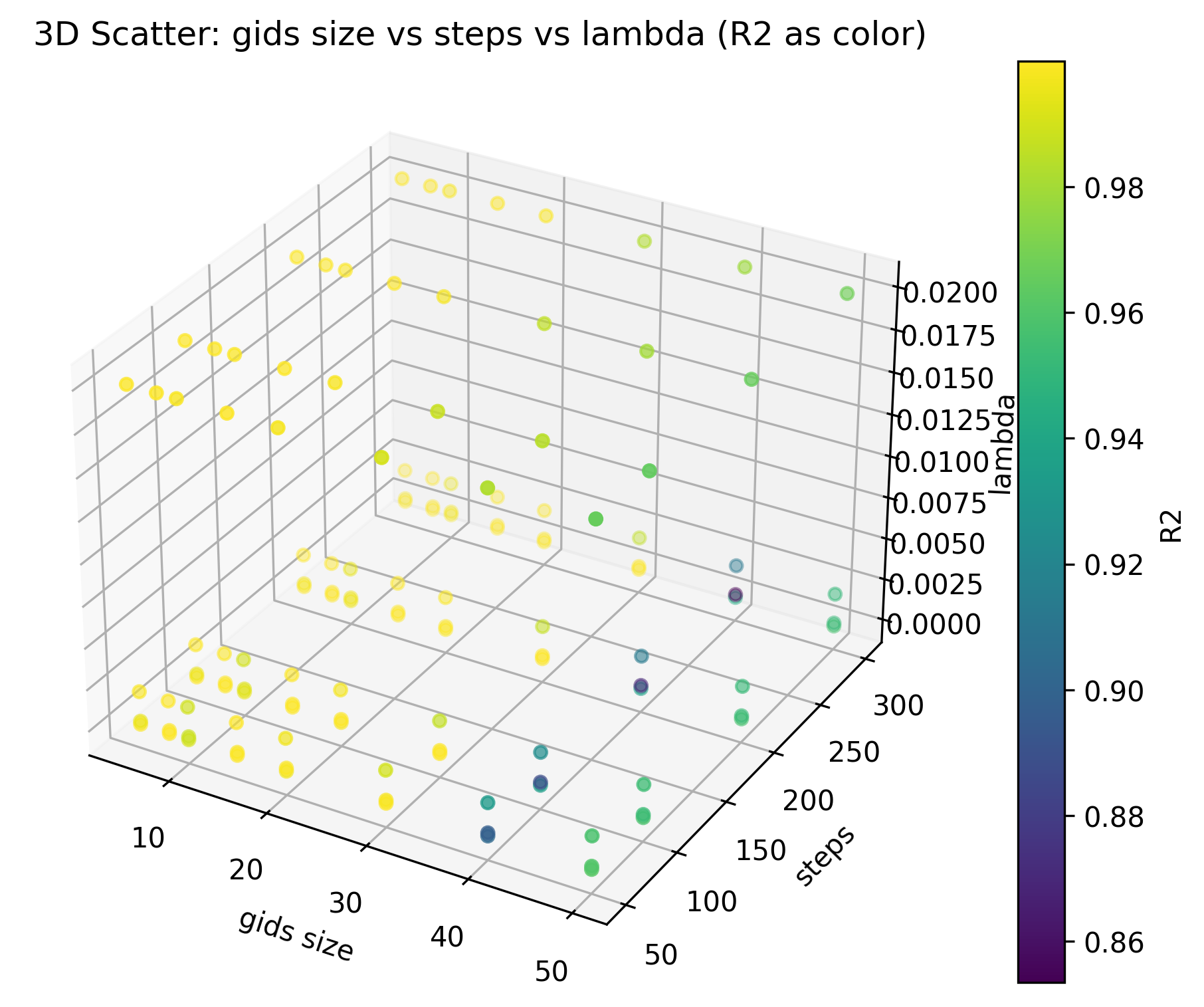} 
        \caption{Parameter search for CI.}
        \label{fig:kan_lambda}
    \end{subfigure}
    
    \caption{KAN model performance as a function of grid size, steps and lambda. The conclusions for indoor and outdoor models are similar to the ABG and CI models.}
    \label{fig:kan_Pareto}
\end{figure*}

\textit{Hyperparameter selection:} To quantify the trade-off between model performance performance and computational complexity, we analyzed the sensitivity of the coefficient of determination ($R^2$) to variations in structural complexity, parameterized by the KAN grid size ($G$), steps and $\lambda$ contextualizing the parameters selected in Table~\ref{tab:KANs_hyperparams}. As illustrated in Figure~\ref{fig:kan_Pareto}, the performance landscape demonstrates that accuracy does not scale linearly with complexity; rather, it exhibits a rapid saturation characteristic. Specifically, for both the ABG and CI datasets, models with minimal grid densities ($G \in [3, 10]$) consistently achieve near-optimal accuracy ($R^2 > 0.98$, indicated by high-intensity markers), comparable to dense models with significantly higher degrees of freedom ($G=50$). This plateau in performance suggests that the underlying physical dynamics can be effectively captured by low-complexity symbolic representations. Considering also the results obtained in our recent work on eCAL, a new metric developed to assess the energy cost of artificial intelligence lifecycle~\cite{11298182}, which shows that with the KAN architectures considered in this paper, together with small training data, AutoPL can run on personal laptops. Moreover, for training, it consumes less than 2 Joules of energy on average.

The final selection of tuned hyperparameters for the PL models learned by KANs are summarized in Table~\ref{tab:KANs_hyperparams}. For both the CI and ABG models, we normalize the training data to maintain consistency with our previous work \cite{anaqreh}. In preliminary experiments without normalization, the frequency, expressed in hertz in the CI model, tended to dominate the other input variables, resulting in imbalanced feature contributions.
\begin{figure*}[h]
    \centering
    \includegraphics[width=1\textwidth]{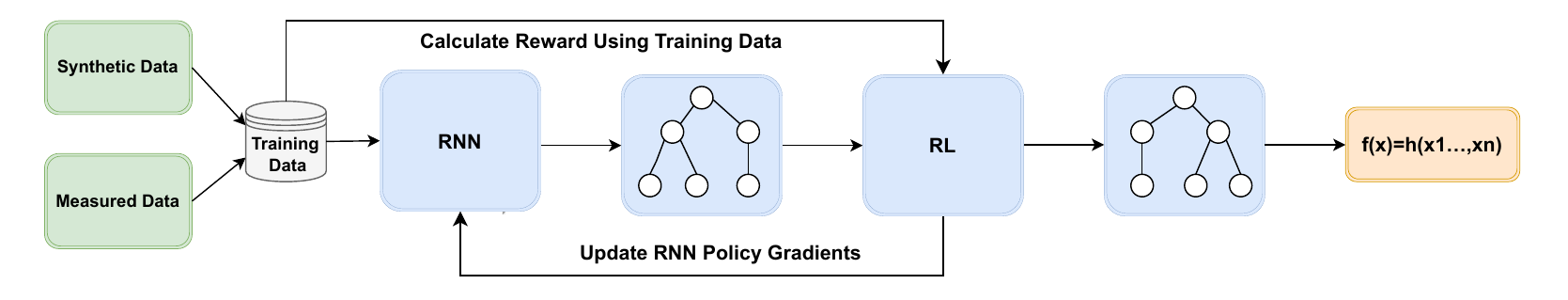} 
    \caption{The flow diagram of developing a PL model using DSR.}
    \label{fig:dsr_workflow}
\end{figure*}
\subsection{Training of DSR}
\label{sec:dsrtr}
\begin{figure*}[h!]
    \centering
    \begin{subfigure}[b]{0.35\textwidth}
        \centering
        \includegraphics[width=\textwidth]{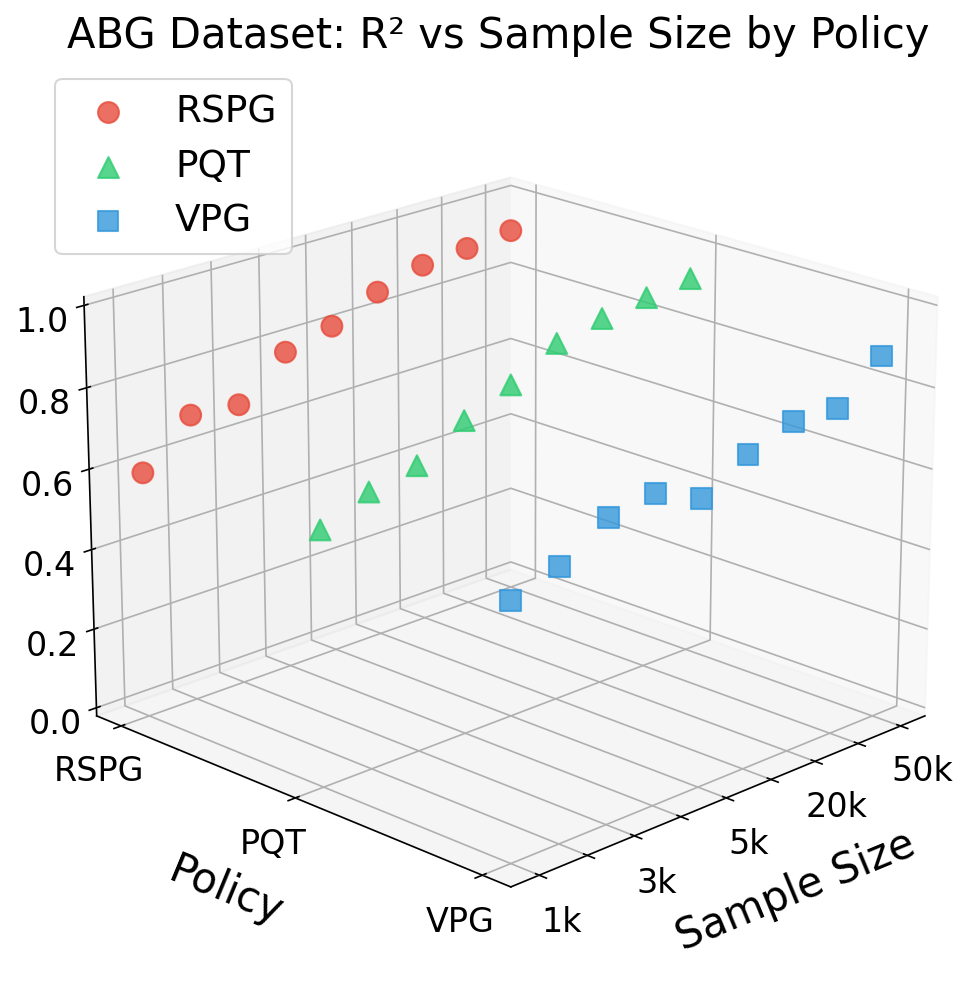}
        \caption{Parameter search for ABG.}
        \label{fig:dsr_grid_r2}
    \end{subfigure}
    \begin{subfigure}[b]{0.35\textwidth}
        \centering
        \includegraphics[width=\textwidth]{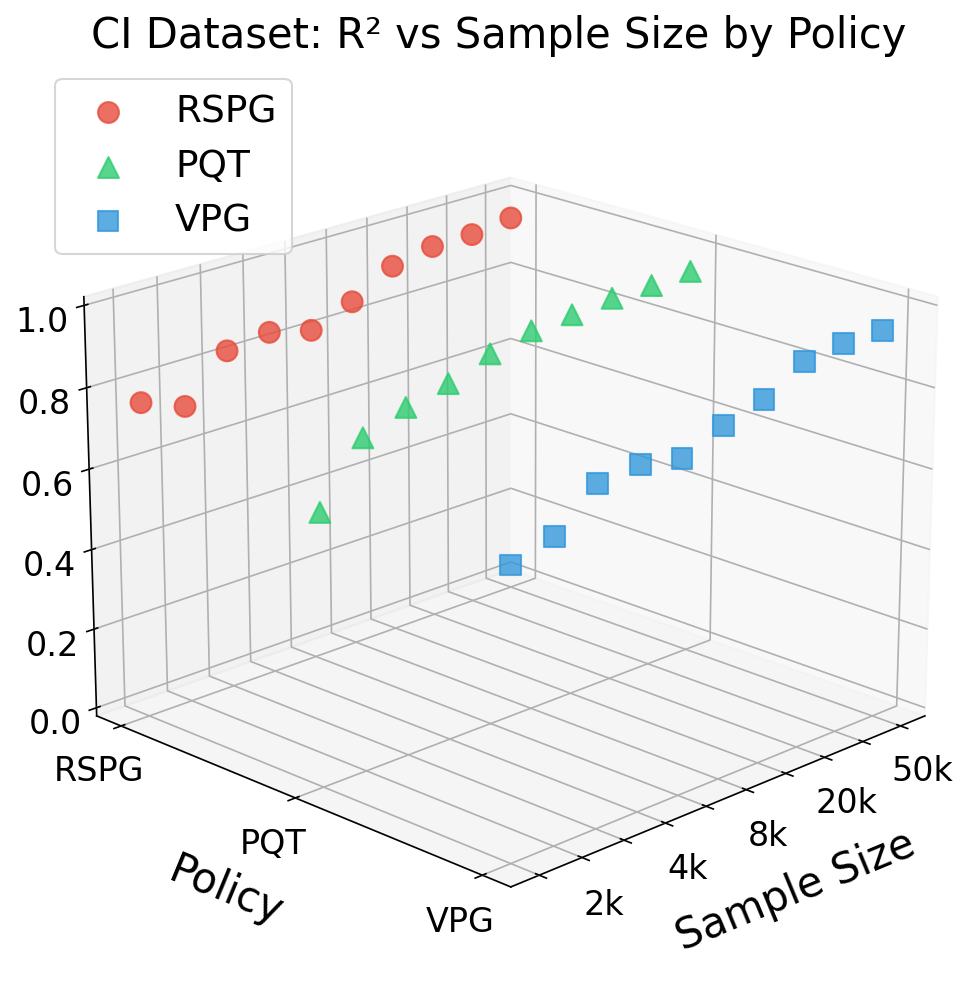} 
        \caption{Parameter search for CI.}
        \label{fig:dsr_lambda}
    \end{subfigure}
    
    \caption{DSR model performance as a function of policy and samples. The conclusions for indoor and outdoor models are similar to the ABG and CI models.}
    \label{fig:dsr_Pareto}
\end{figure*}
We used the DSR implementation provided by the deep\_symbolic\_optimization library\footnote{https://github.com/dso-org/deep-symbolic-optimization}, by training it in a manner consistent with standard machine learning models. Specifically, we employ  80\% of the data for training and 20\% for testing. The flow diagram for DSR method is illustrated in Figure \ref{fig:dsr_workflow}. As depicted in the figure, RNN generates symbolic expressions from the training data while adhering to predefined constraints. RL guides the process by assigning a reward to each expression based on its fit to the training data. Using the policy gradient methods, introduced in Section~\ref{sec:dsr}, the RL algorithm updates the RNN parameters to iteratively improve the generated expressions. This process continues until a stopping criterion, such as convergence or a performance threshold, is met. Finally, the best symbolic expression is obtained.

In DSR, a set of constraints can be implemented to narrow the search space, ensuring the generation of meaningful expressions while preserving computational efficiency. The operator library was selected to align with the functional forms commonly found in established PL models (logarithmic, polynomial, and linear operators), while the constraints described below ensure physically meaningful and interpretable outputs. The following constraints were specifically applied. \textbf{Expression Length Limits}: Expressions were constrained to a minimum length of 4 tokens to avoid trivial solutions and a maximum length of 40 tokens to maintain interpretability. \textbf{Avoiding Redundant Constants:} Operators were restricted to prevent all their children from being constants, preventing the generation of expressions that simplify to a single constant. \textbf{Unary Operator Consistency:} Unary operators were restricted from having their inverse as a child (e.g., \( \log(\exp(x)) \)) to prevent the generation of redundant expressions. \textbf{Trigonometric Operator Composition:} Trigonometric operators were restricted from having other trigonometric operators as descendants (e.g., \( \sin(1 + \cos(x)) \)) to prevent unnecessary complexity, despite being mathematically valid. \textbf{Repeat Constraint:} This constraint regulates the occurrence of specific tokens (e.g., operators and/or variables) in generated expressions by enforcing minimum and maximum limits. It softly penalizes violations of the minimum threshold to allow exploration while strictly invalidating expressions that exceed the maximum limit. The repeat constraint with $f$ and $d$ was utilized in all cases.

\textit{Hyperparameter selection:}  To quantify the trade-off between model performance and computational complexity, we analyzed the sensitivity of the coefficient of determination ($R^2$) to variations in structural complexity, which is shown in Figure~\ref{fig:dsr_Pareto}. This figure is parameterized by the DSR policy and sample size contextualizing the parameters selected in Table~\ref{tab:DSR_hyperparams}. In addition, the 3D scatter plots reveal distinct patterns in how the three DSR-based policies perform across varying sample sizes. For the ABG dataset, all three policies start at similar $R^2$ values (0.59-0.61) with only 1k samples, but their performance trajectories diverge significantly as sample size increases. RSPG and PQT maintain comparable performance throughout, reaching $R^2$ values of 0.92 and 0.93, respectively, at 50k samples, while VPG consistently lags behind, achieving only 0.68 at 5k and 0.88 at 50k samples. This suggests that risk-seeking and priority queue training strategies are more sample-efficient for the ABG model structure. For the CI dataset, an interesting crossover pattern emerges: RSPG achieves the highest $R^2$ (0.76) at 1k samples, but PQT surpasses both alternatives at medium sample sizes (0.90 at 5k vs 0.82 for RSPG). At larger sample sizes (20k-50k), all three policies converge to high performance ($R^{2}>0.93$), with RSPG slightly leading at 50k (0.95). The CI dataset appears more amenable to symbolic regression overall, with all policies achieving higher $R^{2}$ values compared to ABG at equivalent sample sizes. Both figures demonstrate that increasing sample size yields diminishing returns beyond 20k-30k samples, where performance begins to plateau across all policies.       

Each of the three RNN-based algorithms (RSPG, PQT, and VPG) has one unique algorithm-specific parameter, for which we adopted default values as presented in \cite{Petersen}, i.e., the risk factor ($\epsilon = 0.05$) for RSPG, the priority queue size ($k = 10$) for PQT, and the EWMA coefficient ($\alpha = 0.25$) for VPG. In addition to these unique parameters, all three algorithms share a common set of hyperparameters (batch size, learning rate, and entropy weight). We explored various values for these hyperparameters to achieve the best performance. The final tuned hyperparameters are summarized in Table~\ref{tab:DSR_hyperparams}.

\begin{table}[h!]
\centering
\caption{Tuned hyperparameters for RNN-based algorithms in DSR.}
\begin{tabular}{|c|c|c|c|}
\hline
\textbf{Parameter}  & \textbf{RSPG} & \textbf{PQT} & \textbf{VPG} \\\hline
\multicolumn{4}{|c|}{ABG model}\\
\hline
Samples & 50k & 20k & 30k \\
Batch size  & 200 & 200 & 200 \\
Learning rate  & 0.002 & 0.002 & 0.0001 \\
Entropy weight  & 0.008 & 0.005 & 0.005 \\
\hline
\multicolumn{4}{|c|}{CI model}\\\hline
Samples & 2k & 3k & 1k\\
Batch size  & 200 & 200 & 200 \\
Learning rate  & 0.001 & 0.002 & 0.0005 \\
Entropy weight  & 0.008 & 0.005 & 0.008 \\
\hline
\multicolumn{4}{|c|}{Indoor model}\\\hline
Samples & 50k & 50k & 50k \\
Batch size  & 300 & 200 & 200 \\
Learning rate  & 0.0005 & 0.001 & 0.001 \\
Entropy weight  & 0.03 & 0.01 & 0.02 \\
\hline
\multicolumn{4}{|c|}{Outdoor model}\\\hline
Samples & 50k & 50k & 50k \\
Batch size  & 200 & 200 & 200 \\
Learning rate  & 0.0005 & 0.0005 & 0.0001 \\
Entropy weight  & 0.01 & 0.01 & 0.01 \\
\hline
\end{tabular}
\label{tab:DSR_hyperparams}
\end{table}

\begin{table*}[h!]
\centering
\caption{Performance comparison on ABG model. The interpretability and validity assessment follow the methodology in Section \ref{sec:metrics}.}
\label{tab:abg_performance_comparison}

\setlength{\tabcolsep}{3pt} 
\small 

\begin{tabular}{|m{2.5cm}| 
                >{\centering\arraybackslash}m{1.3cm}| 
                >{\centering\arraybackslash}m{1.3cm}| 
                >{\centering\arraybackslash}m{1.3cm}| 
                >{\centering\arraybackslash}m{1.3cm}| 
                m{5.5cm}| 
                >{\centering\arraybackslash}m{1.8cm}| 
                >{\centering\arraybackslash}m{1.2cm}|}
\hline

\diagbox{\textbf{Methods}}{\textbf{Metrics}} & 
\textbf{MAE} \par $\pm$\textbf{STD} & 
\textbf{MSE} \par $\pm$\textbf{STD} & 
\textbf{MAPE} \par $\pm$\textbf{STD} & 
\textbf{R\textsuperscript{2}} \par $\pm$\textbf{STD} & 
\centering \textbf{Expression} & 
\textbf{Interpret.} & 
\textbf{Validity} \\ \hline \hline

\textbf{KAN [6,6,1]*} & 
0.39 \par $\pm$0.74 & 
1.27 \par $\pm$3.35 & 
2.86 \par $\pm$3.63 & 
0.99 \par $\pm$0.00 & 
\centering Tree and splines in Fig.~\ref{fig:ABGkan_network} & 
Medium & ? \\ \hline

\textbf{KAN [6,6,1]*} \par \textbf{Auto-symbolic} & 
3.50 \par $\pm$1.60 & 
25.19 \par $\pm$17.70 & 
22.79 \par $\pm$25.67 & 
0.94 \par $\pm$0.04 & 
\centering $\displaystyle - 291.5 \log_{10}{(8.7 - 3.1 \alpha)} + 32.2 \cos{(1.1 \gamma - 1.3)} + 9 \beta - 53.3 \log_{10}{(9.4 - 3.9 d)} + 0.02 (0.03 f - 1)^{2} + 313.3 - \frac{0.5}{(-f - 0.2)^{2}} + 8 \chi_\sigma^{\text{ABG}}$ & 
High & \cmark \\ \hline \hline

\textbf{DSR-RSPG} &
4.57 \par $\pm$0.25 &
33.26 \par $\pm$3.45 &
11.86 \par $\pm$0.86 &
0.92 \par $\pm$0.01 & 
\centering $20\alpha+10\gamma+\beta+\log_{10}(d)+\frac{\gamma f}{10}+\chi_\sigma^{\text{ABG}}$ & 
High & \cmark \\ \hline

\textbf{DSR-PQT} &
5.07 \par $\pm$0.48 &
42.09 \par $\pm$7.06 &
12.89 \par $\pm$1.65 &
0.90 \par $\pm$0.02 & 
\centering $20\alpha+15\gamma+\beta+\frac{f}{20}+\chi_\sigma^{\text{ABG}}$ & 
High & \xmark \\ \hline


\textbf{DSR-VPG} &
7.43 \par $\pm$1.11 &
87.38 \par $\pm$23.52 &
19.12 \par $\pm$2.91 &
0.79 \par $\pm$0.06 &
\centering $(\alpha+\gamma)(10+\chi_\sigma^{\text{ABG}})$ & 
High & \xmark \\ \hline \hline

\textbf{ResNet-MLP} & 
1.68 \par $\pm$0.24 & 
4.91 \par $\pm$1.45 & 
15.63 \par $\pm$21.99 & 
0.99 \par $\pm$0.00 & 
\centering / & 
none & \xmark \\ \hline

\textbf{TabNet} & 
1.61 \par $\pm$0.35 & 
5.39 \par $\pm$2.66 & 
12.77 \par $\pm$15.48 & 
0.99 \par $\pm$0.00 & 
\centering Feature importance: $\alpha$: 0.3908, $\gamma$: 0.1914, $\beta$: 0.0812, $f$: 0.0614, $d$: 0.2029, $\chi_\sigma^{\text{ABG}}$: 0.0723 & 
Low & \xmark \\ \hline

\end{tabular}
\begin{flushleft}
\footnotesize
* Input parameters were normalized by dividing each value by the maximum value of its corresponding parameter
\end{flushleft}
\end{table*}

\subsection{Evaluation Metrics}
\label{sec:metrics}
The performance of the modeling methods was evaluated using four key metrics: Mean Absolute Error (MAE), Mean Squared Error (MSE), Mean Absolute Percentage Error (MAPE), and the coefficient of determination (\(R^2\)). MAE measures the average absolute difference between the predicted and actual values. MSE emphasizes larger errors by squaring the deviations. MAPE expresses prediction accuracy as a percentage of actual values, providing a normalized measure that allows for meaningful comparison. Finally, \(R^2\) quantifies the proportion of variance in the actual values by indicating how well the predictions capture the variability present in the actual values.

To assess the interpretability of the resulting model, we consider the three levels defined in Section \ref{sec:met}, namely low, medium, and high. Finally, to assess the validity of the expression, we perform a human expert evaluation, according to the flow depicted in Figure~\ref{fig:modeling_workflow}. We explicitly examine the final symbolic expressions produced by both DSR and KAN Auto-symbolic to verify whether they preserve the expected monotonic increase of PL with respect to distance and frequency. Expressions containing oscillatory terms (such as sine or cosine of distance or frequency) that violate this fundamental large-scale propagation property are therefore marked as physically invalid in the evaluation tables. Such monotonicity analysis is however, not straightforward or sometimes even possible with other black-box AI models.

\subsection{DL-based baseline}

To establish robust DL baselines for the PL datasets, we implemented two SotA architectures: a ResNet-MLP~\cite{Gorishniy3540261.3541708} and TabNet~\cite{arik2021tabnet}. The ResNet-MLP adapts residual learning for tabular data, featuring an input projection followed by two residual blocks with a hidden dimension of 64, with each block incorporates batch normalization, ReLU activation, and a dropout rate of 0.1 to mitigate overfitting. We also employed TabNet, an attentive interpretable tabular learning architecture, configured with sequential attention. Both models were trained for 150 epochs with a batch size of 32 using the Adam optimizers with learning rates of $10^{-3}$ and $10^{-2}$, respectively. Similar to the proposed methods, to ensure statistical reliability, performance was evaluated using a Monte Carlo cross-validation protocol with 10 independent runs, utilizing an 80/20 train-test split and standard scaling for feature normalization.

\section{Evaluation Results}
\label{sec:results}
In this section, we evaluate the efficiency of the methods introduced in Section \ref{sec:met} to address the problem outlined in Section \ref{sec:probstat}, using the methodologies described in Section \ref{sec:methodology}. We provide a discussion with a summary of the findings at the end of the section.

\subsection{Learning the ABG Model from Synthetic Data}
\begin{figure}[h!]
    \centering
    \includegraphics[width=0.5\textwidth]{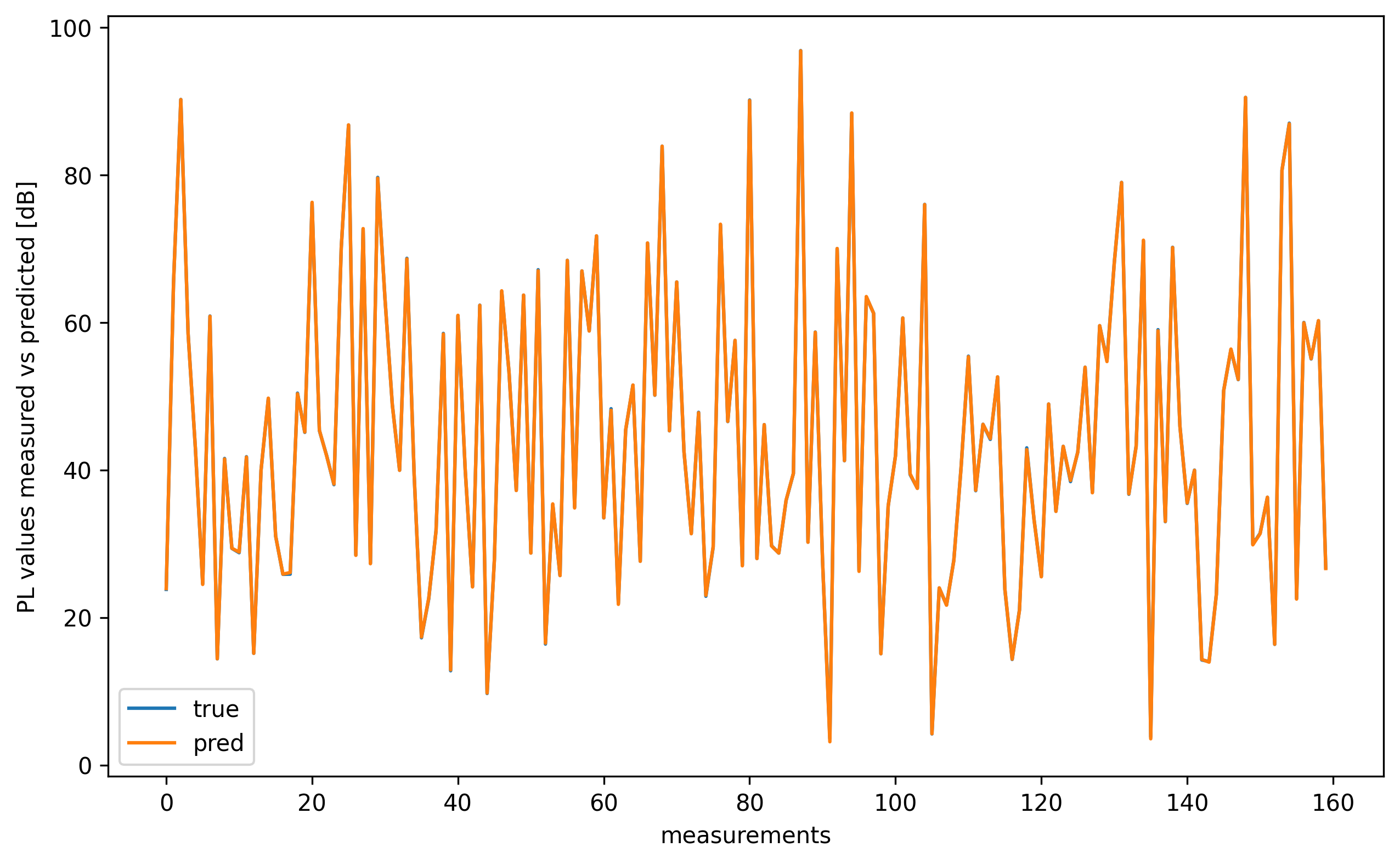} 
    \caption{The predicted values by the KAN model vs the actual generated by $PL^{ABG}$ according to Eq.~(\ref{eq:abg}) as per parameters in Table~\ref{tab:ABG_CI_parameters}.}
    \label{fig:kan_model}
\end{figure}

The performance comparison for the KAN and DSR methods on $PL^{ABG}$ is presented in Table \ref{tab:abg_performance_comparison}. The KAN [6,6,1] achieves superior predictive performance, with exceptionally low error metrics (MAE = 0.39$\pm$0.74, MSE = 1.27$\pm$3.35, MAPE = 2.86$\pm$3.63) and $R^2$ value of 0.99$\pm$0.00, indicating a nearly perfect alignment between the model predictions and the actual data as confirmed by the plot in Figure~\ref{fig:kan_model}. In the figure,  \textit{pred} stands for the predicted PL values and \textit{true} stands for the actual values based on Eq.~(\ref{eq:abg}). In contrast, the KAN [6,6,1] Auto-symbolic, which replaces the learned splines with known functions and re-learns new weights for the edges of the graph, shows significantly lower accuracy, with MAE = 3.50$\pm$1.60, MSE = 25.19$\pm$17.70, MAPE = 22.79$\pm$25.67, and $R^2$ = 0.94$\pm$0.04.

Among DSR models, DSR-RSPG obtained the best performance (MAE = 4.57 $\pm$0.25, MSE = 33.26 $\pm$3.45, MAPE = 11.86 $\pm$0.86  and $R^{2}$ = 0.92 $\pm$0.01), followed by DSR-PQT (MAE = 5.07 $\pm$0.48, MSE = 42.09 $\pm$7.06, MAPE = 12.89 $\pm$1.65, $R^{2}$ = 0.90 $\pm$0.02) and DSR-VPG (MAE = 7.43 $\pm$1.11, MSE = 87.38 $\pm$23.52, MAPE = 19.12 $\pm$2.91, $R^{2}$ = 0.79 $\pm$0.06). While the DSR models may show improvements at high training samples seizes, the training times are significantly larger, taking even days,  compared with KANs  where the most complex took hours. 

Comparing to the DL baselines, ResNet-MLP and TabNet, we can see that both models achieved high predictive stability with $R^2$ scores of $0.99\pm0.00$, comparable to the best KAN. However, their error metrics were notably higher than the KAN [6,6,1], with ResNet-MLP yielding an MAE of 1.68$\pm$0.24 and TabNet an MAE of 1.61$\pm$0.35. These results highlight that while KAN [6,6,1] achieves the highest accuracy, other methods show varying performance metrics, generally with higher error rates and lower values of $R^2$.

\begin{figure}[h!]
    \centering
    \includegraphics[width=0.5\textwidth]{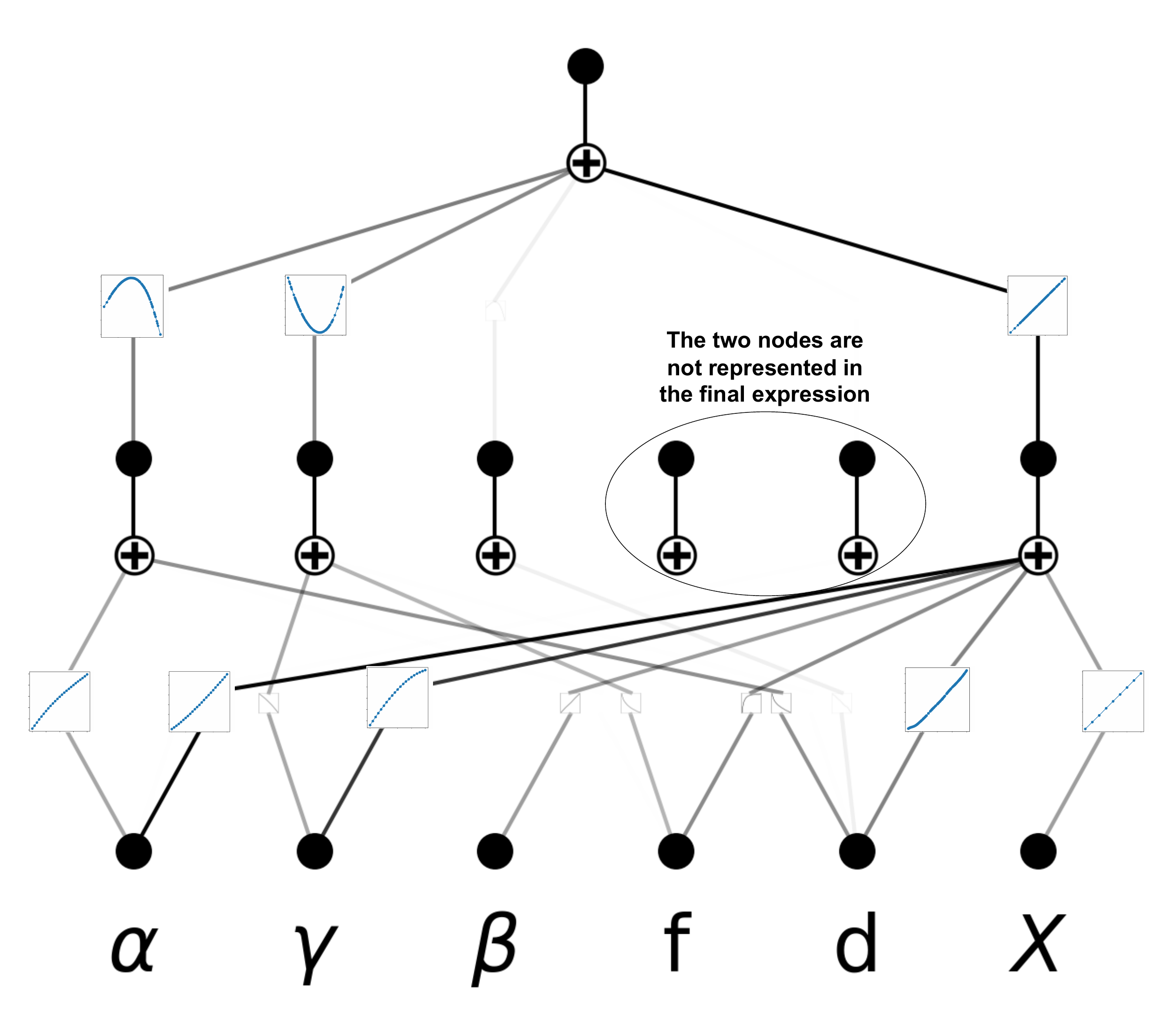} 
    \caption{The learned KAN [6,6,1] network for approximating $PL^{ABG}$.}
    \label{fig:ABGkan_network}
\end{figure}



\paragraph*{Interpretability} As discussed in Sections \ref{sec:kan} and \ref{sec:kantr}, KANs learn splines and weights on the edges, and the best configuration is found using grid search. In our experiments, the best KAN architecture learned for $PL^{ABG}$ is depicted in Figure \ref{fig:ABGkan_network} and has a three-layer design [6,6,1]. The first layer includes six input nodes, corresponding to the parameters $\alpha, \gamma, \beta, f, d, \chi_{\sigma}$; the second layer has the same number of nodes, and the output layer consists of a single node representing the predicted value of the function. 


It can be seen from the learned graph in Figure \ref{fig:ABGkan_network} that the final expression is the summation of four terms, with the third one having a lower weight, i.e. very light gray coloring of the edge that connects the hidden layer with the output layer. The fourth and fifth edges from the last layer were learned as being unimportant, resulting in the disconnection of their associated summation nodes from the overall composition graph. It can also be observed from the functions learned on the first three edges of the last layer of the graph that they perform a nonlinear transformation, while the last one performs a linear one. From the coloring of the edges and summations in the hidden layer, the first summation node, a quasi-linear transformation of $\alpha$ is summed with a nonlinear transformation of $d$ with medium weighting. In the second summation node, a medium weighted quasi-linear transformation of $\gamma$ with a nonlinear transformation of $f$. The next three summation nodes have low to minimal weighting, while the last summation node aggregates all six inputs with high to medium weights. 

Looking at the DL baselines, neither produces a symbolic expression. ResNet-MLP remains a complete black box, while TabNet provides only low-level interpretability via feature importance scores ($\alpha \approx 0.39$, $d \approx 0.20$, and $\gamma \approx 0.19$) without revealing the explicit functional relationships governing the PL.

\paragraph*{Validity} The validity labels reported in Tables V–VIII explicitly account for monotonicity violations with respect to distance and frequency as discussed in Section \ref{sec:metrics}. We examine the validity of the expressions generated by the KANs and DSR methods. For $PL^{ABG}$, the expressions obtained by KAN [6,6,1] Auto-symbolic and DSR-RSPG are considered valid, as they correctly represent PL as a function of $f$ and $d$. We also notice that the expression learned by the KAN [6,6,1] Auto-symbolic is more complex than the one learned by the DSR-RSPG. Further methodological tools are needed as future work to assess the validity of the composition graph and spline from Figure \ref{fig:ABGkan_network}. In contrast, the expression generated by DSR-PQT does not include $d$, and the expression of DSR-VPG excludes both $f$ and $d$. These results indicate that the DSR-PQT and DSR-VPG expressions are unlikely to be valid to accurately model PL from a communication engineering perspective.



\begin{table*}[h!]
\centering
\caption{Performance comparison on CI model. The interpretability and validity assessment follow the methodology in Section \ref{sec:metrics}.}
\label{tab:ci_performance_comparison}

\setlength{\tabcolsep}{3pt} 
\small 

\begin{tabular}{|m{2.5cm}| 
                >{\centering\arraybackslash}m{1.3cm}| 
                >{\centering\arraybackslash}m{1.3cm}| 
                >{\centering\arraybackslash}m{1.3cm}| 
                >{\centering\arraybackslash}m{1.3cm}| 
                m{5.5cm}| 
                >{\centering\arraybackslash}m{1.8cm}| 
                >{\centering\arraybackslash}m{1.2cm}|}
\hline

\diagbox{\textbf{Methods}}{\textbf{Metrics}} & 
\textbf{MAE} \par $\pm$\textbf{STD} & 
\textbf{MSE} \par $\pm$\textbf{STD} & 
\textbf{MAPE} \par $\pm$\textbf{STD} & 
\textbf{R\textsuperscript{2}} \par $\pm$\textbf{STD} & 
\centering \textbf{Expression} & 
\textbf{Interpret.} & 
\textbf{Validity} \\ \hline \hline

\textbf{KAN [4,4,1]*} & 
0.37 \par $\pm$0.22 & 
1.32 \par $\pm$1.21 & 
0.29 \par $\pm$0.16 & 
1.00 \par $\pm$0.00 & 
\centering Tree and splines in Figure \ref{fig:CIkan_network} & 
Medium & ? \\ \hline

\textbf{KAN [4,4,1]*} \par \textbf{Auto-symbolic} & 
3.39 \par $\pm$2.12 & 
24.55 \par $\pm$24.41 & 
2.27 \par $\pm$1.39 & 
0.98 \par $\pm$0.02 & 
\centering $\displaystyle - 280.5 \log_{10}{(8.7 - 1.6 n)} + 43.8 \log_{10}{(1.1 d + 1.9)} + 19.9 \log_{10}{(5.2 f + 9.3)} + 50.8 \log_{10}{(0.8 \chi_\sigma^{\text{CI}} + 7.5)} + 352.2$ & 
High & \cmark \\ \hline \hline

\textbf{DSR-RSPG} &
13.15 \par $\pm$3.26 &
296.77 \par $\pm$137.93 &
8.63 \par $\pm$2.05 &
0.72 \par $\pm$0.13 &

\centering $23n+\frac{d}{10}+\log_{10}(f)+40$ & 
High & \cmark \\ \hline

\textbf{DSR-PQT} &
10.40 \par $\pm$2.69 &
178.09 \par $\pm$72.00 &
6.81 \par $\pm$1.74 &
0.83 \par $\pm$0.07 &

\centering $23n+\frac{d}{10}+(\chi_\sigma^{\text{CI}}+n)\cos(\log_{10}(f))+40$ & 
High & \xmark \\ \hline

\textbf{DSR-VPG} &
14.52 \par $\pm$2.08 &
336.70 \par $\pm$98.19 &
9.52 \par $\pm$1.38 &
0.68 \par $\pm$0.09 &

\centering $21n+2\chi_\sigma^{\text{CI}}+60$ & 
High & \xmark \\ \hline \hline

\textbf{ResNet-MLP} & 
3.59 \par $\pm$0.71 & 
32.99 \par $\pm$12.35 & 
2.56 \par $\pm$0.54 & 
0.97 \par $\pm$0.01 & 
\centering / & 
None & \xmark \\ \hline

\textbf{TabNet} & 
2.21 \par $\pm$0.41 & 
11.44 \par $\pm$4.97 & 
1.54 \par $\pm$0.31 & 
0.99 \par $\pm$0.00 & 
\centering Feature importance: $f$: 0.2891, $n$: 0.3825, $d$: 0.2136, $\chi_\sigma^{\text{CI}}$: 0.1147 & 
Low & \xmark \\ \hline

\end{tabular}

\begin{flushleft}
\scriptsize
* Input parameters were normalized by dividing each value by the maximum value of its corresponding parameter
\end{flushleft}
\end{table*}

\subsection{Learning the CI Model from Synthetic Data}
\begin{figure}[h!]
    \centering
    \includegraphics[width=0.5\textwidth]{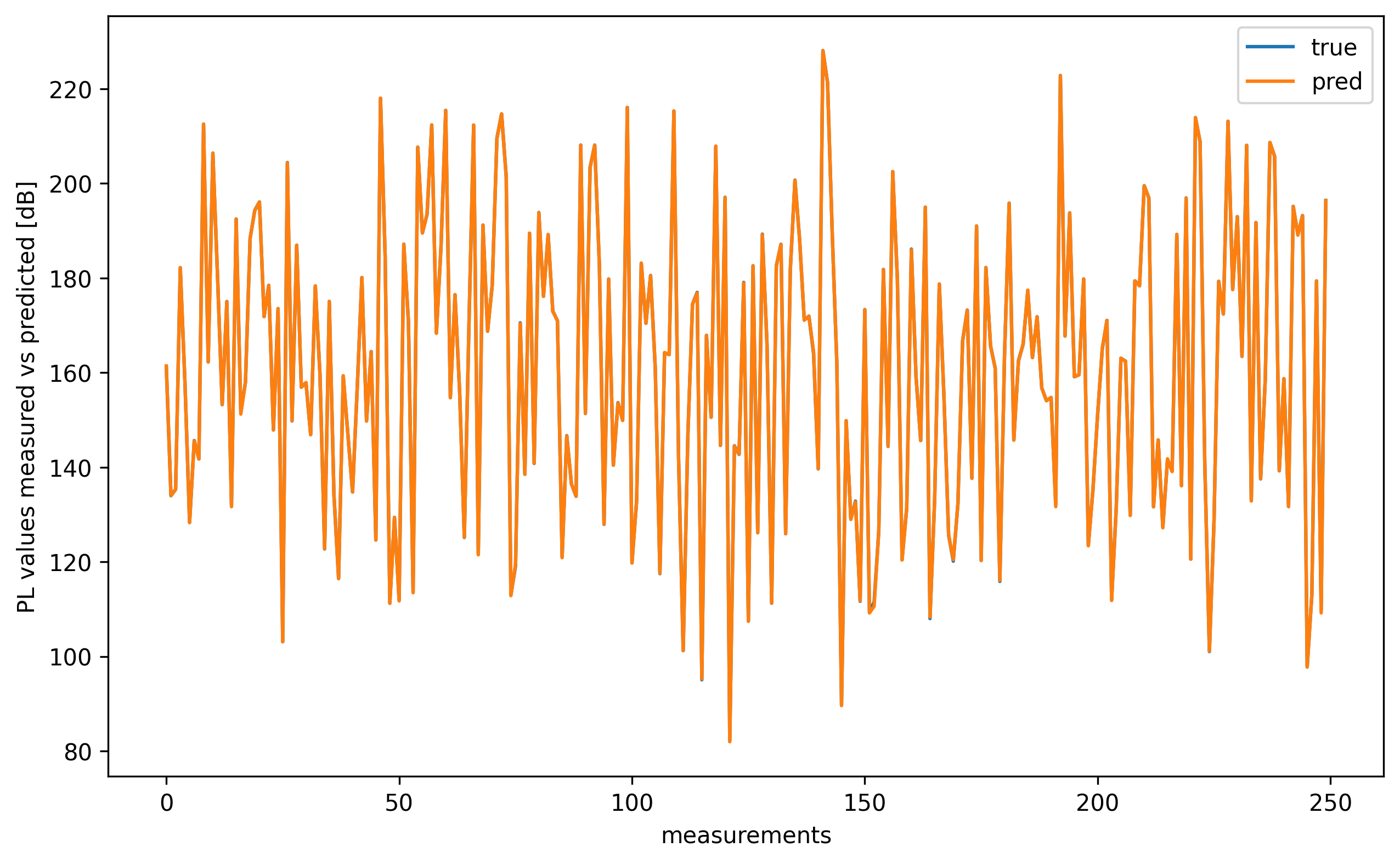} 
    \caption{The predicted values by the KAN model vs the actual generated by $PL^{CI}$ according to Eq.~(\ref {eq:ci}) as per parameters in Table~\ref{tab:ABG_CI_parameters}.}
    \label{fig:CIkan_model}
\end{figure}

Table~\ref{tab:ci_performance_comparison} presents the performance comparison of various methods for discovering $PL^{CI}$. The KAN [4,4,1] method achieved the best results, with minimal error values (MAE = 0.37$\pm$0.22, MSE = 1.32$\pm$1.21, MAPE = 0.29$\pm$0.16) and an $R^2$ value of 1.00$\pm$0.00, showing a highly accurate fit between the predicted values and the actual values as illustrated in Figure~\ref{fig:CIkan_model}. In the figure, \textit{pred} represents the predicted PL values and \textit{true} represents the actual values based on Eq.~(\ref{eq:ci}).
The KAN [4,4,1] Auto-symbolic showed lower performance, with MAE = 3.39$\pm$2.12, MSE = 24.55$\pm$24.41, MAPE = 2.27$\pm$1.39, and $R^2$ = 0.98$\pm$0.02. 

\begin{figure}[h!]
    \centering
    \includegraphics[width=0.5\textwidth]{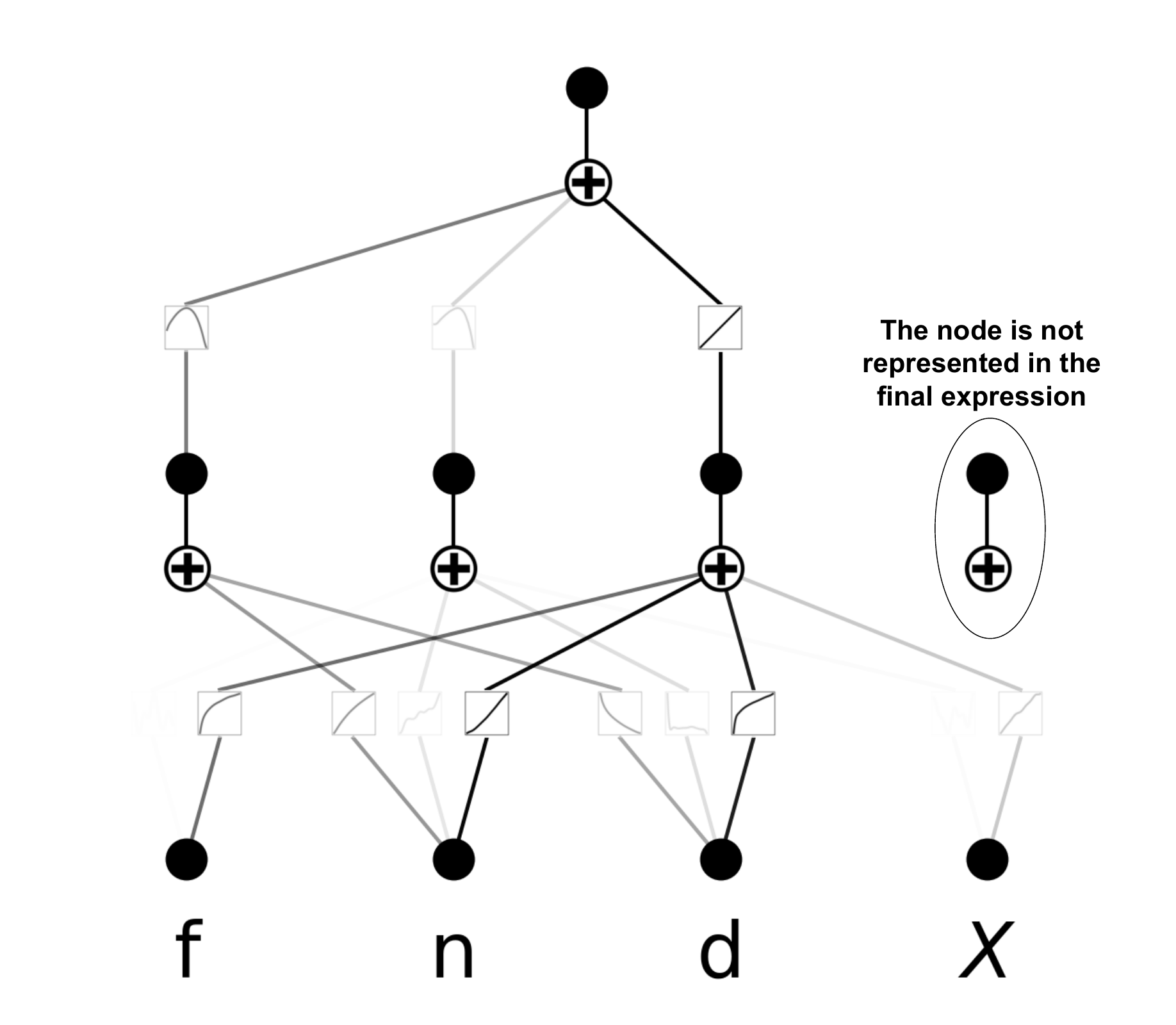} 
    \caption{The learned KAN [4,4,1] network for approximating $PL^{CI}$.}
    \label{fig:CIkan_network}
\end{figure}

Among the DSR methods, DSR-PQT  yielded the lowest error (MAE = 10.40 $\pm$2.69, MSE = 178.09 $\pm$72.00, MAPE = 6.81 $\pm$1.74, $R^{2}$ = 0.83 $\pm$0.07) followed by DSR-RSPG (MAE = 13.15$\pm$3.26, MSE = 296.77$\pm$137.93, MAPE=8.63$\pm$2.05, $R^{2}$=0.72$\pm$0.13) and DSR-VPG (MAE = 14.52 $\pm$2.08, MSE = 336.70 $\pm$98.19, MAPE = 9.52 $\pm$1.38, $R^{2}$ = 0.68 $\pm$0.09). Unlike with the DSR models for ABG, with CI we were able to find more interpretable expressions with significantly faster training time at an order of magnitude lower samples as can be seen in Table \ref{tab:DSR_hyperparams}. As we increase the sample numbers, all four performance metrics improve, however the expressions become more complex.

Comparing to the DL baselines for the CI dataset, TabNet demonstrated superior predictive stability with an $R^2$ of $0.99\pm0.00$, whereas ResNet-MLP achieved a slightly lower $R^2$ of $0.97\pm0.01$. In terms of error metrics, TabNet significantly outperformed the ResNet-MLP, yielding an MAE of 2.21$\pm$0.41, MSE of 11.44$\pm$4.97, and MAPE of 1.54$\pm$0.31, compared to the ResNet-MLP's MAE of 3.59$\pm$0.71, MSE of 32.99$\pm$12.35, and MAPE of 2.56$\pm$0.54. However, both significantly underperformed compared to KAN [4,4,1]. These results demonstrate that the KAN [4,4,1] significantly outperforms the other methods, with smaller errors corresponding to better predictive accuracy.

\paragraph*{Interpretability} Figure~\ref{fig:CIkan_network} shows the KAN architecture utilized for $PL^{CI}$ modeling, featuring a three-layer design [4, 4, 1]. The first layer consists of four input nodes corresponding to the parameters $ f,n, d, \chi_\sigma^{\text{CI}}$, followed by a hidden layer with four nodes, and a single-node output layer representing the predicted function value.


From the learned graph in Figure~\ref{fig:CIkan_network}, it can be seen that the final expression is the sum of three terms, where the second has a lower weight. The rightmost node is disconnected because the model has learned that the connecting edges are not important. The functions associated with the first two edges of the last layer apply nonlinear transformations, whereas the third edge performs a quasi-linear transformation. From the coloring of the edges and summations in the hidden layer, it can be observed that, in the first summation node, a quasi-linear transformation of $n$ and a nonlinear transformation of $d$ are combined, both with medium weighting. The second summation node aggregates the nonlinear transformation of $n$ and $d$ with low weights. Finally, the third summation node aggregates all four inputs with high to medium weights. Looking at the DL baselines, only TabNet provides a low-level interpretability via feature importance scores (f $\approx $ 0.29, n $\approx $  0.38, d $\approx $  0.21, $\chi_\sigma^{\text{CI}} \approx$ 0.11) without revealing the explicit functional relationships to the PL.

\paragraph*{Validity} The expressions generated by KAN and DSRs for the $PL^{CI}$ demonstrate the validity of those modeled by KAN [4,4,1] Auto-symbolic and DSR-RSPG, as in both cases the PL is appropriately modeled as a function of $f$ and $d$. We note that, same as in the case of the $PL^{ABG}$, the expression produced by the KAN [4,4,1] Auto-symbolic is quite complex, while the composition graph requires additional investigation. The expression generated by DSR-PQT, although it includes both $f$ and $d$, models the frequency term using $\cos(\log_{10}(f))$. This implies a periodic contribution of $f$ to PL, which contradicts the fundamental characteristics of signal propagation, as PL increases with increasing $f$. Furthermore, the expression obtained by DSR-VPG excludes both $f$ and $d$. These results imply the invalidity of the DSR-PQT and DSR-VPG expressions for the PL modeling.

\begin{table*}[h!]
\centering
\caption{Performance comparison in an indoor environment. The interpretability and validity assessment follow the methodology in Section \ref{sec:metrics}.}
\label{tab:indoor_performance_comparison}

\setlength{\tabcolsep}{3pt} 
\small 

\begin{tabular}{|m{2.8cm}| 
                >{\centering\arraybackslash}m{1.3cm}| 
                >{\centering\arraybackslash}m{1.3cm}| 
                >{\centering\arraybackslash}m{1.3cm}| 
                >{\centering\arraybackslash}m{1.3cm}| 
                m{5.5cm}| 
                >{\centering\arraybackslash}m{1.8cm}| 
                >{\centering\arraybackslash}m{1.2cm}|}
\hline

\diagbox{\textbf{Methods}}{\textbf{Metrics}} & 
\textbf{MAE} \par $\pm$\textbf{STD} & 
\textbf{MSE} \par $\pm$\textbf{STD} & 
\textbf{MAPE} \par $\pm$\textbf{STD} & 
\textbf{R\textsuperscript{2}} \par $\pm$\textbf{STD} & 
\centering \textbf{Expression} & 
\textbf{Interpret.} & 
\textbf{Validity} \\ \hline \hline

$\boldsymbol{PL^{MWF}}$ Eq. \ref{eq:fw}* & 
9.01 & 121.54 & 8.98 & 0.56 & 
\centering $10n \log_{10}\left(d\right) + PL_0 + \mathrm{WAF} + \mathrm{FAF}$ & 
High & \cmark \\ \hline

$\boldsymbol{PL^{EI}}$ Eq. \ref{eq:in}* & 
7.64 & 111.67 & 7.52 & 0.49 & 
\centering $10n \log_{10}(d) + PL_0 + n_w L_w + n_f^{\left( \frac{n_f + 2}{n_f + 1} - b \right)} L_f$ & 
High & \cmark \\ \hline \hline

\textbf{KAN [4,1]} & 
5.89 \par $\pm$0.13 & 
57.67 \par $\pm$3.30 & 
5.84 \par $\pm$0.12 & 
0.77 \par $\pm$0.02 & 
\centering Learned tree and splines as in Figure \ref{fig:Indoorkan_network} & 
Medium & ? \\ \hline

\textbf{KAN [4,1]} \par \textbf{Auto-symbolic} & 
6.33 \par $\pm$0.23 & 
63.51 \par $\pm$5.41 & 
6.33 \par $\pm$0.21 & 
0.74 \par $\pm$0.02 & 
\centering $41.5 \log_{10}\left(0.4 d + 4.03 \right)- 1.3 \cos\left(5.04 f - 0.5 \right)$ \par $+ 4.9 \sin\left(8.2 n_w - 2.9 \right)+52.03 \log_{10}\left(2.4 n_f + 8.6 \right)- 11.2$ & 
High & \xmark \\ \hline \hline

\textbf{DSR-RSPG} &
6.58 \par $\pm$0.23 &
67.85 \par $\pm$1.90 &
6.69 \par $\pm$0.29 &
0.73 \par $\pm$0.01 &

\centering $\frac{d}{2}+3\log_{10}(f)+5n_w+70$ & 
High & \cmark \\ \hline

\textbf{DSR-PQT} &
6.48 \par $\pm$0.10 &
66.11 \par $\pm$2.07 &
6.59 \par $\pm$0.14 &
0.74 \par $\pm$0.01 &

\centering $\frac{d}{2}+4n_w+80$ & 
High & \xmark \\ \hline

\textbf{DSR-VPG} &
6.39 \par $\pm$0.12 &
64.25 \par $\pm$2.39 &
6.48 \par $\pm$0.18 &
0.75 \par $\pm$0.01 &

\centering $n_f^2+4n_f+n_fn_w+80$ & 
High & \xmark \\ \hline \hline

\textbf{ResNet-MLP} & 
5.67 \par $\pm$0.38 & 
54.97 \par $\pm$7.71 & 
5.54 \par $\pm$0.36 & 
0.78 \par $\pm$0.03 & 
\centering / & 
None & \xmark \\ \hline

\textbf{TabNet} & 
5.14 \par $\pm$0.27 & 
44.02 \par $\pm$5.01 & 
5.09 \par $\pm$0.28 & 
0.82 \par $\pm$0.02 & 
\centering Feature importance: $n_w$: 0.1332, $n_f$: 0.2643, $d$: 0.5505, $f$: 0.0520 & 
Low & \xmark \\ \hline

\end{tabular}

\begin{flushleft}
\scriptsize
* Expressions developed with traditional methods used as baselines.
\end{flushleft}
\end{table*}

\subsection{Learning an Indoor Model from Empirical Data}
\begin{figure}[h!]
    \centering
    \includegraphics[width=0.5\textwidth]{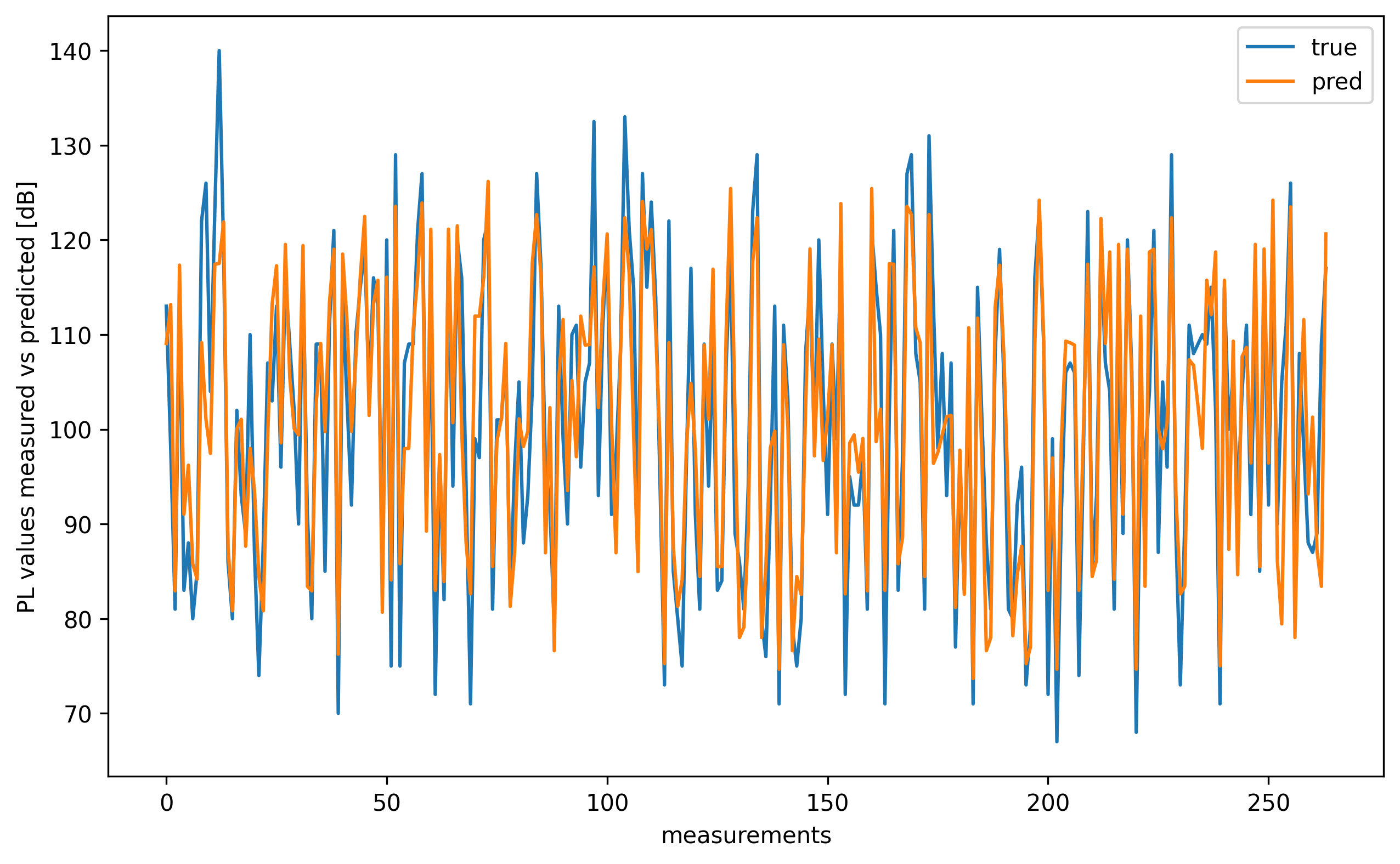} 
    \caption{The predicted values by the KAN model vs the actual indoor measurements.} 
    \label{fig:IndoorTruevsPred}
\end{figure}

The performance comparison of various methods for modeling PL in an indoor environment is presented in Table~\ref{tab:indoor_performance_comparison}. As $PL^{MWF}$ and $PL^{EI}$ are deterministic analytical formulas, their performance metrics are reported as fixed values (STD = 0). The $PL^{MWF}$ serves as the baseline, yielding the highest errors (MAE = 9.01, MSE = 121.54, MAPE = 8.98) and $R^2$ value of 0.56, indicating a moderate predictive performance. $PL^{EI}$, based on Eq.~(\ref{eq:in}) from \cite{ElChall}, slightly improves accuracy, with MAE = 7.64, MSE = 111.67, MAPE = 7.52, and $R^2$ = 0.49. However, all automated methods outperform both the baseline and the empirical model.  KAN [4,1] significantly improves accuracy (MAE = 5.89$\pm$0.13, MSE = 57.67$\pm$3.30, MAPE = 5.84$\pm$0.12) and a substantially higher $R^2$ value of 0.77$\pm$0.02, as shown in Figure~\ref{fig:IndoorTruevsPred}, where \textit{pred} stands for the predicted PL values and \textit{true} stands for the actual measurements from \cite{ElChall}. The KAN [4,1] Auto-Symbolic performs similarly, with slightly higher errors but maintaining the same $R^2$ value. 

\begin{figure}[h!]
    \centering
    \includegraphics[width=0.3\textwidth]{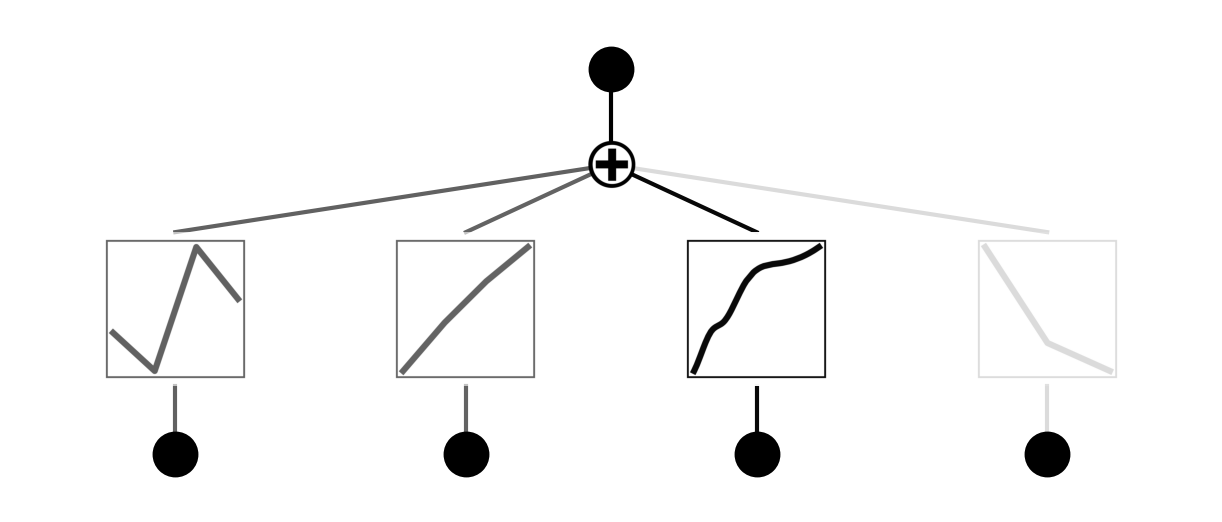} 
    \caption{The learned KAN [4,1] network for approximating $PL^{EI}$.}
    \label{fig:Indoorkan_network}
\end{figure}

Among DSR methods, DSR-VPG is slightly superior to DSR-PQT that also slightly surpasses DSR-RSPG. DRR-VPG achieves (MAE = .39 $\pm$0.12, MSE = 64.25 $\pm$2.39, MAPE = 6.48 $\pm$0.18 and $R^{2}$ = 0.75 $\pm$0.01), though its $R^2$ remains inferior to KAN and slightly surpasses KAN auto-symbolic. DSR-PQT and DSR-RSPG demonstrate comparatively weaker performance, with slightly higher errors and lower values of $R^2$ (0.74$\pm$0.01 and 0.73 $\pm$0.01, respectively). 

In the indoor environment, the DL baselines demonstrated significantly higher predictive accuracy compared to the analytical models ($PL^{MWF}$ and $PL^{EI}$). TabNet achieved the best performance among the methods with an $R^2$ of 0.82$\pm$0.02, an MAE of 5.14$\pm$0.27, MSE of 44.02$\pm$5.01, and MAPE of 5.09$\pm$0.28. The ResNet-MLP followed with a slightly lower $R^2$ of 0.78$\pm$0.03, MAE of 5.67$\pm$0.38, MSE of 54.97$\pm$7.71, and MAPE of 5.54$\pm$0.36.
Overall, the results highlight that both KANs and DSR models provide substantial improvements over the indoor empirical models, both slightly underperformed compared to the DL baselines, but lack interpretability.

\paragraph*{Interpretability} Figure~\ref{fig:Indoorkan_network} illustrates the architecture of the KAN [4,1] used for indoor PL prediction. The architecture has two layers; the first layer consists of four input nodes corresponding to the parameters $ n_w,n_f, d, f$, and a single-node output layer representing the predicted value. As illustrated in the figure, the final expression is a sum of four terms, with the fourth term contributing less due to its smaller weight. Notably, the function learned on the first edge applies a nonlinear transformation, whereas the remaining three edges implement quasi-linear transformations. Despite their superior accuracy, neither DL baseline offers a transparent functional form. ResNet-MLP acts as a black box, while TabNet provides limited interpretability through feature importance ($d \approx 0.55$, $n_f \approx 0.26$, $n_w \approx 0.13$, $f \approx 0.05$).

\paragraph*{Validity} For the indoor environment, the KAN [4,1] Auto-symbolic expression contains a $\cos(f)$ term, implying a periodic dependence on frequency. Physically, PL represents the average signal attenuation caused by wavefront spreading and interactions with the environment~\cite{physical_PL_avg} (e.g., reflection, diffraction, and material penetration). These phenomena make the attenuation typically increase monotonically with frequency and distance. Therefore, oscillatory forms like cosine or sine functions are characteristics of small-scale fading rather than large-scale fading, i.e., PL discussed in this work, are inappropriate to be used for modeling PL. To prevent such anomalies, physics-guided constraints should be applied, such as limiting the search space to monotonic operators.
In contrast, the expression obtained by DSR-RSPG is valid, as it includes both $f$ and $d$. Furthermore, the findings highlight the invalidity of the expressions of PQT and VPG: the expression of DSR-PQT does not include $f$, while the expression of DSR-VPG excludes both $f$ and $d$.
\begin{table*}[h!]
\centering
\caption{Performance comparison in an outdoor environment. The interpretability and validity assessment follow the methodology in Section \ref{sec:metrics}.}
\label{tab:outdoor_performance_comparison}

\setlength{\tabcolsep}{3pt} 
\small 

\begin{tabular}{|m{2.8cm}| 
                >{\centering\arraybackslash}m{1.3cm}| 
                >{\centering\arraybackslash}m{1.3cm}| 
                >{\centering\arraybackslash}m{1.3cm}| 
                >{\centering\arraybackslash}m{1.3cm}| 
                m{5.5cm}| 
                >{\centering\arraybackslash}m{1.8cm}| 
                >{\centering\arraybackslash}m{1.2cm}|}
\hline

\diagbox{\textbf{Methods}}{\textbf{Metrics}} & 
\textbf{MAE} \par $\pm$\textbf{STD} & 
\textbf{MSE} \par $\pm$\textbf{STD} & 
\textbf{MAPE} \par $\pm$\textbf{STD} & 
\textbf{R\textsuperscript{2}} \par $\pm$\textbf{STD} & 
\centering \textbf{Expression} & 
\textbf{Interpret.} & 
\textbf{Validity} \\ \hline \hline

$\boldsymbol{PL^{FS}}$ Eq. \ref{eq:fs}* & 
34.08 & 1242.99 & 31.43 & 0.30 & 
\centering $20 \log_{10}(d) +20 \log_{10}(f) + 32.44$ & 
High & \cmark \\ \hline

$\boldsymbol{PL^{EO}}$ Eq. \ref{eq:out}* & 
18.09 & 398.88 & 17.82 & 0.30 & 
\centering $10n \log_{10}(d) + PL_0 + L_h \log_{10}(h_{\text{ED}}) + X_{\sigma}$ & 
High & \cmark \\ \hline \hline

\textbf{KAN [3,1]} & 
4.83 \par $\pm$0.36 & 
50.54 \par $\pm$12.78 & 
4.66 \par $\pm$0.37 & 
0.57 \par $\pm$0.12 & 
\centering Learned tree and splines as in Figure \ref{fig:Outdoorkan_network} & 
Medium & ? \\ \hline

\textbf{KAN [3,1]} \par \textbf{Auto-symbolic} & 
7.73 \par $\pm$0.35 & 
99.86 \par $\pm$6.05 & 
7.53 \par $\pm$0.43 & 
0.15 \par $\pm$0.04 & 
\centering $\cos{(8.7 f - 3.5)} - 2.3 \log_{10}{(7.7 - 2.2 h_{\text{ED}})} + 105.8$ & 
High & \xmark \\ \hline \hline

\textbf{DSR-RSPG} &
7.69 \par $\pm$0.24 &
95.60 \par $\pm$6.70 &
7.52 \par $\pm$0.26 &
0.28 \par $\pm$0.05 &

\centering $h_{\text{ED}}d +\log_{10}(f)+80$ & 
High & \cmark \\ \hline

\textbf{DSR-PQT} &
7.38 \par $\pm$0.19 &
89.02 \par $\pm$5.00 &
7.20 \par $\pm$0.21 &
0.33 \par $\pm$0.04 &

\centering $\log_{10}(\frac{d}{h_{\text{ED}}})+\frac{d}{10}+90$ & 
High & \xmark \\ \hline

\textbf{DSR-VPG} &
8.02 \par $\pm$0.24 &
101.66 \par $\pm$5.91 &
7.87 \par $\pm$0.28 &
0.23 \par $\pm$0.04 &

\centering $\frac{d+\sin(f)}{10}+90$ & 
High & \xmark \\ \hline \hline

\textbf{ResNet-MLP} & 
7.87 \par $\pm$0.43 & 
101.74 \par $\pm$9.74 & 
7.48 \par $\pm$0.44 & 
0.14 \par $\pm$0.09 & 
\centering / & 
None & \xmark \\ \hline

\textbf{TabNet} & 
7.44 \par $\pm$0.42 & 
90.69 \par $\pm$8.69 & 
7.21 \par $\pm$0.44 & 
0.23 \par $\pm$0.06 & 
\centering Feature importance: $h_{\text{ED}}$: 0.2566, $d$: 0.5952, $f$: 0.1481 & 
Low & \xmark \\ \hline

\end{tabular}

\begin{flushleft}
\scriptsize
* Expressions developed with traditional methods used as baselines.
\end{flushleft}
\end{table*}
\subsection{Learning an Outdoor Model from Empirical Data}
\begin{figure}[h!]
    \centering
    \includegraphics[width=0.5\textwidth]{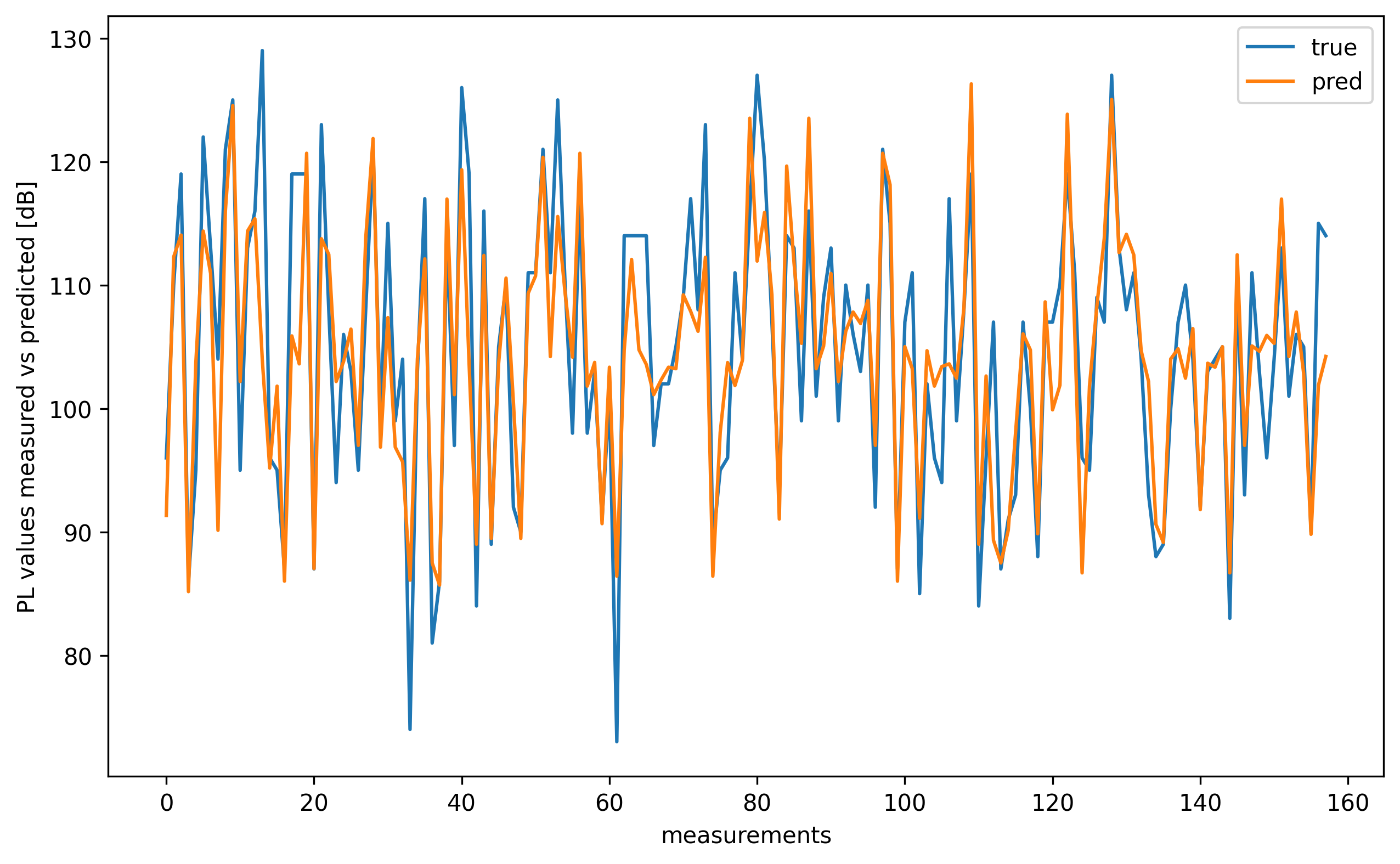} 
    \caption{The predicted values by the KAN model vs the actual outdoor measurements.}
    \label{fig:OutdoorTruevsPred}
\end{figure}

Table~\ref{tab:outdoor_performance_comparison} summarizes the performance of different methods for PL prediction in an outdoor environment. As $PL^{FS}$ and $PL^{EO}$ are deterministic analytical formulas, similar to the indoor dataset, their performance metrics are reported as fixed values (STD = 0). The $PL^{FS}$, yields high error values (MAE = 34.08, MSE = 1242.99, MAPE = 31.43) and a low $R^2$ value of 0.30, indicating poor predictability. The $PL^{EO}$, represented by Eq.~(\ref{eq:out}) from \cite{ElChall}, substantially improves accuracy (MAE = 18.09, MSE = 398.88, MAPE = 17.82), but retains the same $R^2$ value of 0.30. In contrast, the KAN model [3,1] achieves significantly better performance, with the lowest errors (MAE = 4.83$\pm$0.36, MSE = 50.54$\pm$12.78, MAPE = 4.66$\pm$0.37) and a higher $R^2$ value of 0.57$\pm$0.12, as demonstrated in Figure~\ref{fig:OutdoorTruevsPred} showing the comparison between predicted and measured PL values. The KAN [3,1] Auto-symbolic yields poor performance (MAE = 7.73$\pm$0.35, $R^2$ = 0.15$\pm$0.04), showing a higher error and a lower fit compared to the original KAN structure. Furthermore, the very low $R^2$ score shows low interdependence between the input and output variables, meaning that the function to spline matching process resulted in an inaccurate match, leading to significant degradation relative to the KAN-only network. 

Among the DSR models, DSR-PQT performs the best (MAE = 7.38 $\pm$0.19, MSE = 89.02 $\pm$5.00, MAPE = 7.20 $\pm$0.21, $R^{2}$ = 0.33 $\pm$0.04); however, its $R^2$ value remains low. DSR-RSPG and DSR-VPG show slightly higher errors and lower $R^2$ values. 

In the outdoor environment, the DL baselines struggled to capture the underlying PL dynamics effectively compared to the KAN [3,1]. TabNet showed slightly better performance than the ResNet-MLP, achieving an $R^2$ of 0.23$\pm$0.06, MAE of 7.44$\pm$0.42, MSE of 90.69$\pm$8.69, and MAPE of 7.21$\pm$0.44. The ResNet-MLP performed worse, yielding an $R^2$ of only 0.14$\pm$0.09, with an MAE of 7.87$\pm$0.43, MSE of 101.74$\pm$9.74, and MAPE of 7.48$\pm$0.44. Overall, the KAN [3,1] model outperforms both traditional, DSR and DL methods, offering a more accurate method for outdoor PL prediction.

\begin{figure}[h!]
    \centering
    \includegraphics[width=0.4\textwidth]{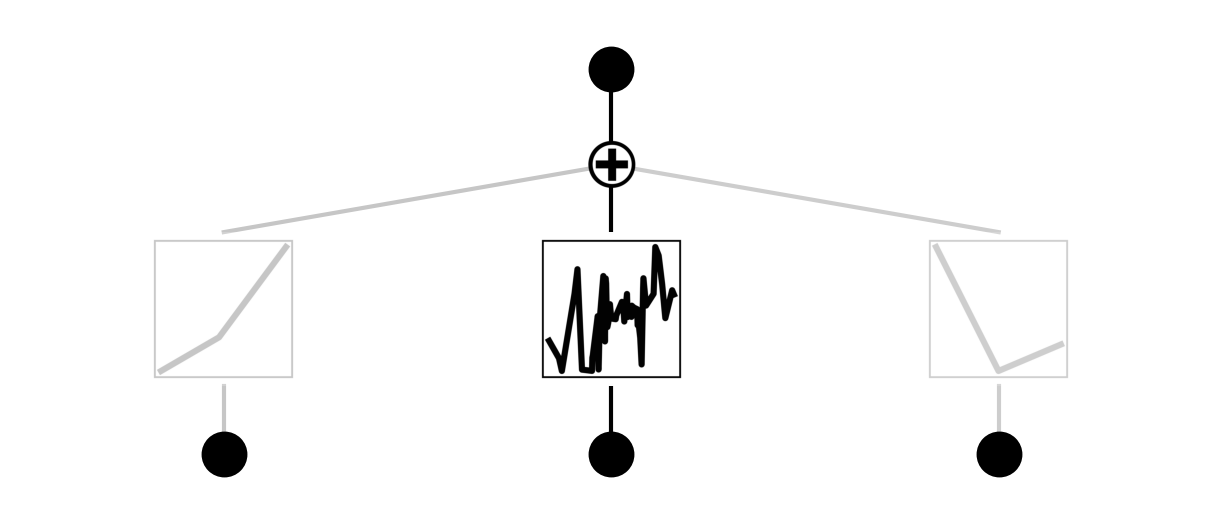} 
    \caption{The learned KAN [3,1] network for approximating $PL^{EO}$.}
    \label{fig:Outdoorkan_network}
\end{figure}

\paragraph*{Interpretability} Figure~\ref{fig:Outdoorkan_network} illustrates the KAN architecture evaluated for the outdoor model, consisting of a two-layer design [3, 1]. The first layer includes three input nodes corresponding to the parameters $h_{\text{ED}}, d, f$, followed by a single output node representing the predicted PL value. The figure shows that the final expression is a sum of three terms, with the first and third terms having smaller weights. It can be observed that the functions learned on the second and third edges apply nonlinear transformations, whereas the first implements a quasi-linear transformation. Regarding DL baselines interpretability, similar to other datasets, ResNet-MLP again functions as a complete black box, whereas TabNet provides limited transparency via feature importance ($d \approx 0.60$, $h_{\text{ED}} \approx 0.26$, $f \approx 0.15$).

\paragraph*{Validity} The only valid expression for the outdoor environment is obtained from DSR-RSPG, as it includes both $f$ and $d$. The KAN [3,1] Auto-symbolic expression excludes $d$ and incorporates a cosine function of $f$, As noted in the indoor case, this introduces a periodic behavior that violates the physics that PL should show a monotonous decay trend over large distances ($d$) and frequencies ($f$).
The expression of DSR-PQT excludes $f$. Finally, although the DSR-VPG expression includes both $f$ and $d$, it defines $f$ using a sine function, introducing a periodicity that misrepresents the actual relationship between $f$ and PL.
\subsection{Discussion}
The evaluation results across ABG, CI, indoor, and outdoor PL modeling scenarios reveal a clear trade-off between the accuracy and the interpretability of expressions generated by KANs and DSR methods.

KANs consistently achieve superior predictive accuracy across all scenarios. For example, KAN [6,6,1] and KAN [4,4,1] achieve exceptionally low error metrics and $R^2$ values of 1, emphasizing their superior ability to model complex data relationships. The learned models, in the form of spline approximations and weights in a graph composition, are more interpretable compared to DL alternatives; however, they are not conventional human-interpretable expressions. To gain insights into the interpretability of the KANs, auto-symbolic is utilized to map from spline-based structures to more interpretable functional forms as illustrated in Figure \ref{fig:kan_workflow} and discussed in Section \ref{sec:kantr}.

Auto-symbolic variants, such as KAN [6,6,1] Auto-symbolic and KAN [3,1] Auto-symbolic, exhibit invalid expressions, higher error rates, and, in some cases, notably lower $R^2$ values compared to their original forms. This illustrates the limits of the current mapping capabilities of KANs, resulting in less accurate and interpretable expressions. In contrast, DSR methods, especially DSR-RSPG, generate simple, valid, and interpretable expressions. Although DSR methods achieve a reasonable level of accuracy, they are computationally more expensive and their predictive performance is generally outperformed by KANs. The lower values observed in the DSR methods indicate a trade-off in which predictive precision is reduced to preserve the simplicity and interpretability of the resulting expressions.

While in this section, the paper analyzed the performance, functional interpretability, and validity of the proposed AutoPL method, with KANs and DSRs while also adding ResNet-MLP and TabNet as alternative neural network baselines, a comprehensive computational cost vs performance Pareto front study is left for future. As AI tools for scientific discovery, including KAN and DSR-based, essentially explore a large search space to find the right distribution for the phenomenon underlying the observed data, the methods themselves are subject to their own hyperparameter settings. To provide insight into this complexity, we considered $R^{2}$ vs selected parameters for both KANs in Section \ref{sec:kantr} and DSR in Section \ref{sec:dsrtr}.

Overall, our findings indicate that KANs deliver highly accurate results with superior first-level interpretability (spline-based structures) compared to typical DNN activation maps. However, when it comes to the second level of interpretability, where actual functions are extracted using the Auto-symbolic, KANs are outperformed by certain DSR methods. On the other hand, DSR methods can be constrained to automatically prioritize interpretability over predictive accuracy while KANs involve slightly more manual labor to achieve that. In our experiments that can be reproduced using the available open source code, we noticed that DSRs tend to take longer to train compared to KANs. The choice between these methods should be determined by the intended objective, whether to maximize predictive accuracy or to derive interpretable expressions that offer clear analytical insight. Generally, the selection and validation of the simplest and most functionally interpretable expression remains manual, as illustrated in Figure \ref{fig:modeling_workflow}, with many potential candidates suggested by AutoPL. Future development to introduce physics constraints during training might speed up also that part.

\subsection{Scope and Limits of Interpretability}
This work emphasizes explicit functional interpretability rather than physical-mechanistic. AutoPL yields closed-form expressions for PL as a function of distance, frequency, wall/floor counts, and antenna height. These expressions expose parameter sensitivities and monotonic trends, enable direct comparison with established models (e.g., CI and ABG), and can be integrated into system-level analysis and planning tools.

The resulting models are not intended to disentangle specific propagation mechanisms, such as reflection, diffraction, scattering. They should therefore be interpreted as empirical, data-driven surrogates, similar in spirit to CI/ABG, that describe what best fits the observed data rather than providing a first-principles electromagnetic account of why propagation behaves as it does.

This aligns with the philosophy of large-scale PL modeling, which favors compact, analytically tractable expressions over physical granularity. Within this framework, AutoPL automates the discovery of such expressions directly from data without imposing predefined model structures.

\subsection{Limitations and Guidelines for Safe Adoption}
\label{subsec:limitations}

While AutoPL provides a powerful framework for the automated discovery of PL models, its deployment in real-world network planning requires an understanding of its inherent limitations. To ensure the safe adoption of the discovered models, we identify the following constraints:

\textit{1) Dataset Scope and Bias:} The accuracy of a discovered symbolic model is strictly bounded by the diversity of the training data. If the input dataset is limited to a specific frequency band, such as sub-6 GHz, or a specific environment (such as Urban Micro), the framework may discover an empirical relationship that captures environment-specific noise rather than a generalizable physical law. Users should ensure that the dataset spans the full operational range of the intended application.

\textit{2) Sensitivity of Auto-Symbolic Mapping:} The transition from a high-dimensional search space (in DSR) or a flexible activation function (in KAN) to a closed-form symbolic expression involves a trade-off between mathematical simplicity and predictive fidelity. There is a risk of "over-simplification," where a discovered model ignores subtle physical inflection points—such as the break-point distance in two-slope models—to achieve a higher "simplicity score." We recommend users inspect the Pareto front of complexity versus accuracy provided by AutoPL before selecting a final model.

\textit{3) Domain Shift and Generalization:} Like most data-driven techniques, models discovered via AutoPL are susceptible to performance degradation when subjected to domain shift. A model discovered in a specific geographical region may fail to generalize to a different city due to variations in building materials, foliage density, or atmospheric conditions. Consequently, a discovered model should be treated as a "specialized" expression for its target domain unless validated against diverse out-of-distribution (OOD) datasets.

While the current AutoPL framework does not enforce physical constraints during optimization, future extensions could incorporate physics-informed constraints directly into the learning objective. Examples include enforcing monotonicity with respect to distance and frequency and restricting the functional search space to non-oscillatory operators. Such constraints could reduce the occurrence of physically implausible expressions while preserving the flexibility of automated model discovery.

\textit{4) Operational Safeguards:} To adopt AutoPL safely, we include a "Human-Scientist-in-the-Loop" validation step as depicted in Figure \ref{fig:modeling_workflow}. 
\section{Conclusion}
\label{sec:con}
The work demonstrates the potential of automated modeling methods, particularly KANs and DSR, to discover accurate and interpretable PL models in wireless communication systems. Through comprehensive evaluation against traditional models such as ABG, CI, and empirical models for outdoor and indoor environments, we demonstrate that KANs consistently achieve superior predictive accuracy across both synthetic and real-world data. Although KANs are less interpretable, they achieve cutting-edge performance. In contrast, DSR generates compact expressions with moderate predictive precision, offering a favorable trade-off between accuracy and interpretability. These results promote the integration of automated methods into the design of next-generation wireless systems, where the demand for efficient and explainable models continues to increase.

This work also revealed several challenges to be addressed in the future. More specifically, utilizing a clear form of PL, the non-logarithmic form, may enable more precise modeling. The performance-interpretability tradeoffs with KANs could likely be improved by enhanced symbolic mapping. Further methodological tools to assess the validity of the composition graph learned by the KAN may ease the adoption of the proposed method. Finally, extending both the KANs and DSR methods to cover diverse scenarios will be essential to validate their robustness and adaptability. These advances have the potential to contribute significantly to the development of intelligent, adaptive modeling frameworks in the field of wireless communication systems.

\textbf{Acknowledgments} This work was supported by the Slovenian Research Agency (P2-0016) and the European Commission NANCY project (No. 101096456).

\bibliographystyle{IEEEtran}
\bibliography{Automated_Modeling_Methods}



\begin{IEEEbiographynophoto}{Ahmad Anaqreh}
received a Master’s degree in Computer Science in 2019 and a Ph.D. in Computer Science in 2024 from the University of Szeged, Hungary. He worked for one year as a postdoctoral researcher at SensorLab, Joszef Stefan Institute, Slovenia. His research focuses on optimizing graph theory problems, particularly NP-hard problems. Currently, he is an assistant professor at Department of Computer Science, Faculty of Information Technology, Applied Science Private University(ASU), Amman, Jordan.
\end{IEEEbiographynophoto}

\begin{IEEEbiographynophoto}{Shih-Kai Chou}
received his Master’s degree in communication engineering from National Chung Cheng University (CCU), Taiwan in 2014 and a Ph.D. from Queen’s University, Belfast, U.K. in 2023. His Ph.D. focused on Reconfigurable Intelligent Surface (RIS/IRS), more specifically, the real-world setup and optimization of RIS-aided wireless systems. currently He is a postdoctoral Researcher with SensorLab, Jožef Stefan Institute.
\end{IEEEbiographynophoto}

\begin{IEEEbiographynophoto}{Bla\v{z} Bertalani\v{c}}
 received the Ph.D. degree (Hons.) from the Faculty of Electrical Engineering, University of Ljubljana. He is currently a Researcher with SensorLab, Jožef Stefan Institute. His research interests include the advancement of machine learning and AI algorithms, especially in the context of time series analysis and smart infrastructures. He held several leadership positions in Slovenian Chapter with over 15 IEEE publications.
\end{IEEEbiographynophoto}

\begin{IEEEbiographynophoto}{Mihael Mohorčič}
is currently the Head of the Department of Communication Systems and a Scientific Counselor with Jožef Stefan Institute, as well as a Full Professor at Jožef Stefan International Postgraduate School. His research interests include advanced wireless communication systems, including mobile, satellite, and stratospheric networks; heterogeneous and ad hoc networks; wireless sensor networks; and the Internet of Things. His current research interests include AI-driven resource management in wireless communications, smart infrastructure connectivity, and intelligent sensing applications. He has contributed to over 25 international research projects in mobile and satellite communications, UAV communication systems, and wireless sensor networks, along with more than 15 national basic and applied research projects. He has co-authored more than 230 journals and conference publications, three books, and one patent. He actively serves on conference organizing committees, including as the general chair and the TPC chair.
\end{IEEEbiographynophoto}

\begin{IEEEbiographynophoto}{Thomas Lagkas}
is an Associate Professor at the Department of Informatics of the Democritus University of Thrace and Director of the Laboratory of Industrial and Educational Embedded Systems. He has been Lecturer and then Senior Lecturer at the Computer Science Department of the International Faculty of The University of Sheffield from 2012 to 2019. He also served as Departmental Research Director and Leader of the ICT Track of the South-East European Research Centre. Furthermore, he has been Adjunct Lecturer at the Department of Informatics and Telecommunications Engineering of the University of Western Macedonia from 2007 to 2013. Moreover, he has been Laboratory and then Scientific Associate at the Technological Educational Institute of Thessaloniki from 2004 to 2012. From 2000 until 2006, he also worked as Programmer undertaking related projects as free lancer.
\end{IEEEbiographynophoto}

\begin{IEEEbiographynophoto}{Carolina Fortuna}
was a Postdoctoral Researcher with Ghent University, Ghent, Belgium. She was a Visiting Researcher at InfoLab, Stanford University, Stanford, CA, USA. She is currently a Research Associate Professor with Jožef Stefan Institute, where she leads SensorLab. She has led and contributed EU-funded projects, such as H2020 NRG5, eWINE, WISHFUL, FP7 CREW, Planetdata, ACTIVE, and under various positions. She has advised/co-advised more than six M.Sc. and Ph.D. students. She has consulted public and private institutions. She has co-authored over 100 articles, including IEEE COMMUNICATIONS SURVEYS AND TUTORIALS, IEEE Wireless Communications Magazine, IEEE OPEN JOURNAL OF THE COMMUNICATIONS SOCIETY, and IEEE ACCESS. Her research interests include developing the next generation of smart infrastructures that surround us and improve the quality of our lives. She contributed to community work as a TPC Member, the Track Chair, and a Reviewer at several IEEE conferences, including Globecom and ICC.
\end{IEEEbiographynophoto}

\vfill
\end{document}